\documentclass[journal]{IEEEtran}
\ifCLASSINFOpdf
\else
\fi

\usepackage{graphicx}
\usepackage{url}
\usepackage{booktabs}
\usepackage{breakurl}
\usepackage{multirow}
\usepackage{array}
\usepackage{tabu}
\usepackage{bbm}
\usepackage{xcolor}
\usepackage{amsmath}
\usepackage{amssymb}
\usepackage{comment}
\usepackage{algorithm}
\usepackage[noend]{algpseudocode}
\usepackage{caption}
\usepackage{tabularx}

\usepackage{float}
\usepackage[caption=false,font=footnotesize]{subfig}

\newcolumntype{C}{>{\centering\arraybackslash}p}
\newcolumntype{L}{>{\raggedright\arraybackslash}p}

\newcolumntype{Z}{>{$}l<{$}}

\usepackage[flushleft]{threeparttable}
\usepackage{tablefootnote}

\newcolumntype{M}{>{$}c<{$}}

\makeatletter
\def\algbackskip{\hskip-\ALG@thistlm}
\makeatother

\usepackage[hang,flushmargin]{footmisc}

\renewcommand{\arraystretch}{1.3}

\makeatletter
\def\footnoterule{\relax%
 \kern-5pt
 \hbox to \columnwidth{\hfill\vrule width \columnwidth height 0.6pt\hfill}
 \kern4.6pt}
\makeatother
\newcolumntype{Y}{>{\centering\arraybackslash}X}

\begin{document}

%
\title{C2CL: Contact to Contactless Fingerprint Matching}
%
%

\author{Steven A. Grosz*, Joshua J. Engelsma, Eryun Liu,~\IEEEmembership{Member,~IEEE} and Anil K. Jain,~\IEEEmembership{Life~Fellow,~IEEE}%
        
}

\twocolumn[{%
\renewcommand\twocolumn[1][]{#1}%
\maketitle
\begin{center}
    \centering
    \footnotesize
    \captionsetup{font=footnotesize}
    \includegraphics[width=1.0\linewidth]{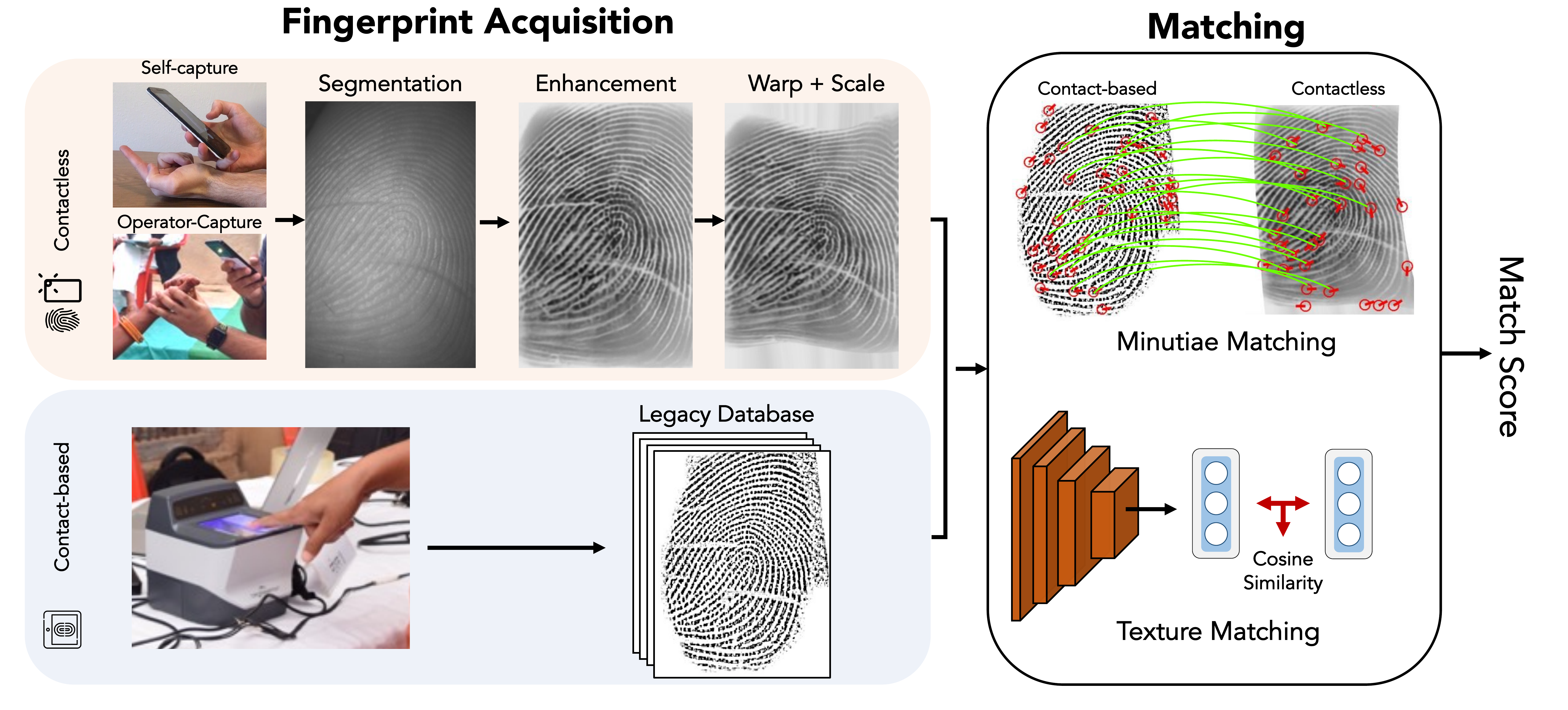}
    \captionof{figure}{\footnotesize Overview of matching contactless fingerprint images with a legacy database of contact-based fingerprint impressions. While only a specific scenario is shown here where contact-based images are obtained from optical FTIR readers (slap or single finger capture) and contactless images are captured by a smartphone camera, our approach can be applied to any heterogeneous fingerprint matching problem.}
    \label{fig:problemOverview}
\end{center}
}]

{
  \renewcommand{\thefootnote}%
    {\fnsymbol{footnote}}
  \footnotetext{
  S.A. Grosz, J.J. Engelsma, and A. K. Jain are with the Department of Computer Science and Engineering, Michigan State University, East Lansing, MI, 48824. E-mail: \{groszste, engelsm7, jain\}@cse.msu.edu \\
  Eryun Liu is with the College of Information Science and Electronic Engineering, Zhejiang University, Hangzhou, China. E-mail: eryunliu@zju.edu.cn \\
  *Corresponding author.
  }
}

\begin{abstract}
Matching contactless fingerprints or finger photos to contact-based fingerprint impressions has received increased attention in the wake of COVID-19 due to the superior hygiene of the contactless acquisition and the widespread availability of low cost mobile phones capable of capturing photos of fingerprints with sufficient resolution for verification purposes. This paper presents an end-to-end automated system, called C2CL, comprised of a mobile finger photo capture app, preprocessing, and matching algorithms to handle the challenges inhibiting previous cross-matching methods; namely i) low ridge-valley contrast of contactless fingerprints, ii) varying roll, pitch, yaw, and distance of the finger to the camera, iii) non-linear distortion of contact-based fingerprints, and vi) different image qualities of smartphone cameras. Our preprocessing algorithm segments, enhances, scales, and unwarps contactless fingerprints, while our matching algorithm extracts both minutiae and texture representations. A sequestered dataset of $9,888$ contactless 2D fingerprints and corresponding contact-based fingerprints from $206$ subjects ($2$ thumbs and $2$ index fingers for each subject) acquired using our mobile capture app is used to evaluate the cross-database performance of our proposed algorithm. Furthermore, additional experimental results on $3$ publicly available datasets show substantial improvement in the state-of-the-art for contact to contactless fingerprint matching (TAR in the range of $96.67\%$ to $98.30\%$ at FAR=$0.01\%$).
\end{abstract}


\IEEEpeerreviewmaketitle

\section{Introduction}

\IEEEPARstart{D}{ue} to their presumed uniqueness and permanence, fingerprints are one of the most widely used biometric traits for secure authentication and search~\cite{pankanti2002individuality, yoon2015longitudinal}. Over the years many different types of fingerprint readers have been developed to obtain a digital image of a finger's friction ridge pattern. These readers vary in a number of different ways, including the underlying sensing technology (\textit{e.g.}, optical, capacitive, ultrasonic, \textit{etc.}) or in the manner in which a user interacts with the reader (\textit{i.e.} contactless, 4-4-2 slap, or single finger contact-based acquisition). Most prevailing fingerprint readers in use today necessitate physical contact of the user's finger with the imaging surface of the reader; however, this direct contact presents certain challenges in processing the acquired fingerprint images. Most notably, elastic human skin introduces a non-linear deformation upon contact with the imaging surface which has been shown to significantly degrade matching performance~\cite{bazen2003fingerprint, cappelli2001modelling, ross2005deformable}.  Furthermore, contact with the surface is likely to leave a latent impression on the imaging surface~\cite{parziale2008touchless}, which presents a security risk as an imposter could illegally gain access to the system though creation of a presentation (\textit{i.e.}, spoof) attack.

In light of the ongoing Covid-19 pandemic, contactless fingerprint recognition has gained renewed interest as a hygienic alternative to contact-based fingerprint acquisition~\cite{okereafor2020fingerprint}. This is further supported by a recent survey that showed that the majority of users prefer touchless capture methods in terms of usability and hygeine considerations~\cite{priesnitz2021mobile}. Prior studies have explored the use of customized 2D or 3D sensing for contactless fingerprint acquisition~\cite{song2004new, lee2006study, parziale2009advanced, hiew2007touch, kumar2011contactless, yin2019contactless}, while others have explored the low-cost alternative of using readily available smartphone cameras to capture ``finger photos''\footnote{In general, contactless fingerprints refers to fingerprint images acquired by a contactless fingerprint sensor, whereas finger photo refers to fingerprint images acquired by a mobile phone. In this paper, we use the two terms interchangeably.}~\cite{sankaran2015smartphone, malhotra2017fingerphoto, stein2012fingerphoto}. 

\begin{figure}
    \centering
    \subfloat[]{\includegraphics[width=1.0\linewidth]{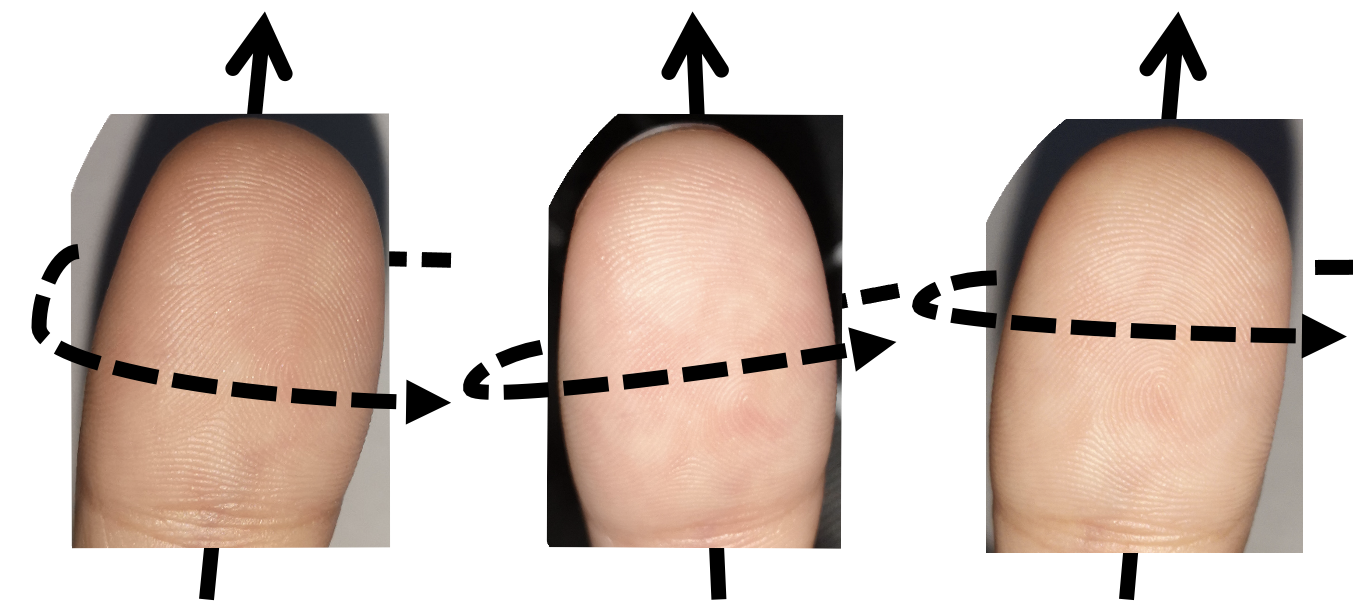}%
    \label{toprow}}\\
    \subfloat[]{\includegraphics[width=1.0\linewidth]{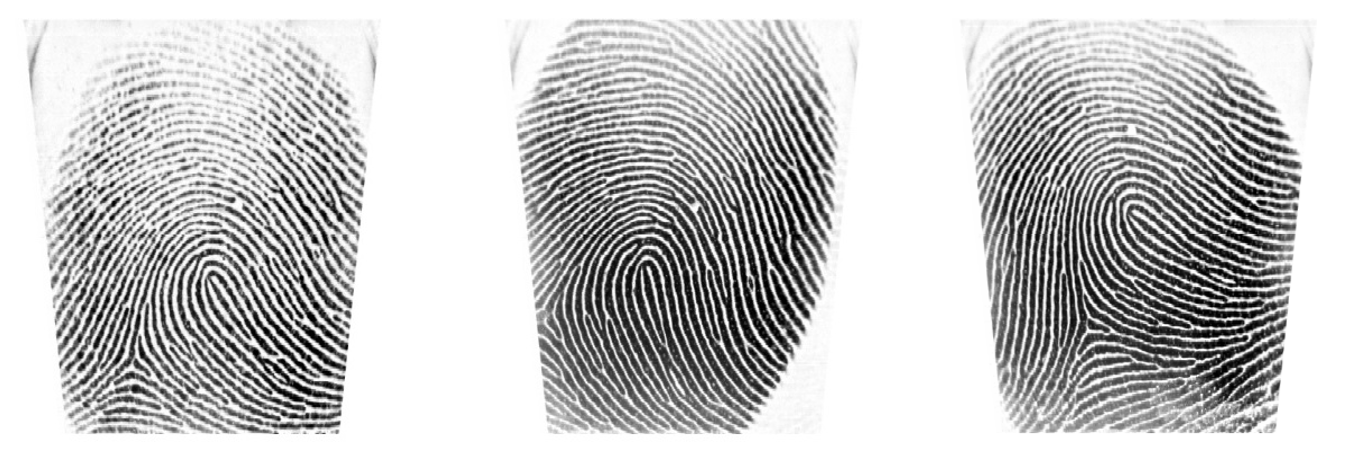}%
    \label{bottomrow}}
    \caption{Examples of contactless fingerprints (a) and their corresponding contact-based fingerprint images (b). Varying viewing angle, resolution, and illumination of contactless images and non-linear distortion of contact-based fingerprints contribute to the degradation of cross-matching performance. The contactless images shown are from the ZJU dataset.}
    \label{fig:varyingViews}
    \vspace{-1.5em}
\end{figure}

Despite the benefits of contactless fingerprint acquisition, imaging and subsequently matching a contactless fingerprint presents its own set of unique challenges. These include (i) low ridge-valley contrast, (ii) non-uniform illumination, (iii) varying roll, pitch, and yaw of the finger, (iv) varying background, (v) perspective distortions due to the varying distances of the finger from the camera, and (vi) lack of cross-compatibility with legacy databases of contact-based fingerprints (see Figure~\ref{fig:varyingViews}). For widespread adoption, contactless fingerprint recognition must overcome the aforementioned challenges and bridge the gap in accuracy compared to contact-contact fingerprint matching.

\begin{figure*}
    \centering
    \subfloat[]{\includegraphics[width=0.165\linewidth]{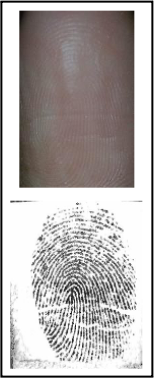}%
    \label{IIT_Bombay_ex_imgs}}
    \hfill
    \subfloat[]{\includegraphics[width=0.165\linewidth]{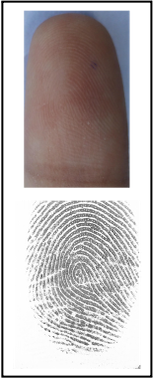}%
    \label{ISPFDv2_ex_imgs}}
    \hfill
    \subfloat[]{\includegraphics[width=0.165\linewidth]{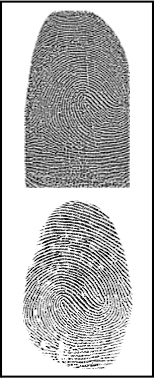}%
    \label{MSU_ex_imgs}}
    \hfill
    \subfloat[]{\includegraphics[width=0.165\linewidth]{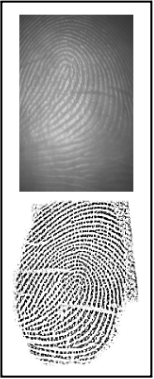}%
    \label{PolyU_ex_imgs}}
    \hfill
    \subfloat[]{\includegraphics[width=0.165\linewidth]{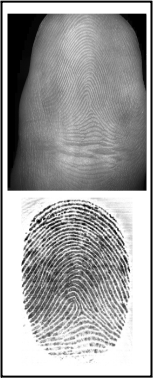}%
    \label{UWA_ex_imgs}}
    \hfill
    \subfloat[]{\includegraphics[width=0.165\linewidth]{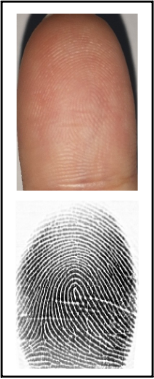}%
    \label{ZJU_ex_imgs}}
    \caption{Example contactless and contact-based fingerprint image pairs from databases which we have obtained from different research groups: (a) IIT Bombay~\cite{birajadar2019towards}, (b) ISPFDv2~\cite{malhotra2020matching}, (c) MSU~\cite{deb2018matching}, (d) PolyU~\cite{lin2018matching}, (e) UWA~\cite{Zhou2014Benchmark}, and (f) ZJU datasets. In general, contactless fingerprints suffer from low ridge-valley contrast, varying roll, pitch, and yaw, and perspective distortions, especially those captured by smartphone cameras (\textit{e.g.}, (a), (b), (c) and (f)). We believe our study involves the largest collection of public domain databases of contactless and contact-based fingerprints.}
    \label{fig:dataset_imgs}
    \vspace{-1.5em}
\end{figure*}


\begin{table*}[t]
\renewcommand{\arraystretch}{1.3}
\caption{Summary of Published Cross-Matching Contact to Contactless Fingerprint Recognition Studies.}
\label{tab:prior_work}
\begin{tabularx}{\linewidth}{l | X | X | L{0.19\textwidth}}
\noalign{\hrule height 1.5pt}
\textbf{Study} & \textbf{Approach} & \textbf{Database} & \textbf{Accuracy}$^\dagger$\\
\noalign{\hrule height 1.0pt}
\multirow{3}{*}{Lin and Kumar, 2018~\cite{lin2018matching}} & \multirow{3}{\hsize}{Robust TPS deformation correction model, minutiae and ridge matching} & $1,800$ contactless and contact fingerprints from $300$ fingers~\cite{lin2018matching}. \newline $2,000$ contactless and $4,000$ contact fingerprints from $1,000$ fingers~\cite{Zhou2014Benchmark} & \multirow{3}{\hsize}{EER = $14.33\%$~\cite{lin2018matching} \newline EER = $19.81\%$~\cite{Zhou2014Benchmark}}\\
\hline
Deb \textit{et al.}, 2018~\cite{deb2018matching} & COTS matcher & $2,472$ contactless and contact fingerprints from $1,236$ fingers~\cite{deb2018matching} & TAR = $92.4\% - 98.6\%$ \newline  @ FAR = $0.1\%$~\cite{deb2018matching}\\
\hline
\multirow{3}{*}{Lin and Kumar, 2019~\cite{lin2019cnn-based}} & \multirow{3}{\hsize}{Fusion of three Siamese CNNs} & $960$ contactless and contact fingerprints from $160$ fingers~\cite{lin2018matching}. \newline $1,000$ contactless and $2,000$ contact fingerprints from $500$ fingers~\cite{Zhou2014Benchmark} & \multirow{3}{\hsize}{EER = $7.93\%$~\cite{lin2018matching} \newline EER = $7.11\%$~\cite{Zhou2014Benchmark}}\\
\hline
Wild \textit{et al.}, 2019~\cite{wild2019comparative} & Filtering based on NFIQ 2.0 quality measure, COTS matcher & $1,728$ contactless and $2,582$ contact fingerprints from $108$ fingers~\cite{wild2019comparative} & TAR = $95.5\% -9 8.6\%$ \newline @ FAR = $0.1\%$~\cite{wild2019comparative}\\
\hline
\multirow{3}{*}{Dabouei \textit{et al.}, 2019~\cite{dabouei2019deep}} & TPS spatial transformer network for deformation correction and binary ridge-map extraction network, COTS matcher & \multirow{3}{\hsize}{$2,000$ contactless and $4,000$ contact fingerprints from $1,000$ fingers~\cite{Zhou2014Benchmark}} & \multirow{3}{\hsize}{EER = $7.71\%$~\cite{Zhou2014Benchmark}}\\
\hline
\multirow{2}{*}{Malhotra \textit{et al.}, 2020~\cite{malhotra2020matching}} & Feature extraction with deep scattering network, random decision forest matcher & \multirow{2}{\hsize}{$8,512$ contactless and $1,216$ contact fingerprints from $152$ fingers~\cite{malhotra2020matching}} & \multirow{2}{\hsize}{EER = $2.11\% - 5.23\%$~\cite{malhotra2020matching}}\\
\hline
\multirow{3}{*}{Priesnitz \textit{et al.}, 2021~\cite{priesnitz2021mobile}} & Neural network-based minutiae feature extraction, open-source minutiae matcher & $896$ contactless from two different capture setups and $464$ contact fingerprints from $232$ fingers~\cite{priesnitz2021mobile} & \multirow{3}{\hsize}{EER = $15.71\%$ and $32.02\%$~\cite{priesnitz2021mobile}}\\
\hline
\multirow{6}{*}{\textbf{Proposed Approach}} & \multirow{6}{\hsize}{TPS spatial transformer for $500$ ppi scaling and deformation correction of contactless fingerprints. Fusion of minutiae and CNN texture representations.} & $8,512$ contactless and $1,216$ contact fingerprints from $152$ fingers~\cite{malhotra2020matching}. \newline $2,000$ contactless and $4,000$ contact fingerprints from $1,000$ fingers~\cite{Zhou2014Benchmark}. \newline $960$ contactless and contact fingerprints from $160$ fingers~\cite{lin2018matching}. \newline $9,888$ contactless and $9,888$ contact fingerprints from $824$ fingers (ZJU Dataset) & \multirow{6}{\hsize}{EER = $1.20\%$~\cite{malhotra2020matching} \newline EER = $0.72\%$~\cite{Zhou2014Benchmark} \newline EER = $0.30\%$~\cite{lin2018matching} \newline EER = $0.62\%$ (ZJU Dataset)}\\
\noalign{\hrule height 1.5pt}
\multicolumn{4}{l}{$\dagger$ Some studies only report EER while other studies only report TAR @ FAR = $0.1\%$.}
\end{tabularx}
\vspace{-1.5em}
\end{table*}

The most significant factor limiting the adoption of contactless fingerprint technology is cross-compatibility with legacy databases of contact-based fingerprints, which is particularly important for governmental agencies and large-scale national ID programs such as India’s Aadhaar National ID program which has already enrolled over 1 billion users based upon contact-based fingerprints. Several studies have aimed at improving the compatibility of matching legacy slap images to contactless fingerprint images~\cite{lin2018matching, deb2018matching, wild2019comparative, dabouei2019deep, lin2019cnn-based, malhotra2020matching}; however, none have achieved the same levels of accuracy as state-of-the-art (SOTA) contact-contact fingerprint matching (such as the results reported in FVC-ongoing~\cite{dorizzi2009fingerprint} and NIST FpVTE~\cite{nist_fpvte}). Furthermore, all of these works focus on solving only a subset of the challenges in an effort to push the contact-contactless matching accuracy closer to the SOTA contact-contact matching systems. Indeed, to the best of our knowledge, this study presents the most comprehensive, end-to-end\footnote{The Cambridge Dictionary defines end-to-end as ``from the very beginning of a process to the very end''. Our method is end-to-end as it carries out the full process from data collection to recognition. Our use of ``end-to-end'' should not be confused with ``end-to-end learning''.} solution in the open academic literature for contact-contactless fingerprint matching that addresses the challenges inherent to each step in the contact to contactless matching process (mobile capture, segmentation, enhancement, scaling, non-linear warping, representation extraction, and matching).

We show that our end-to-end matcher, called C2CL, is able to significantly improve contact-contactless matching performance over the prevailing SOTA methods through experimental results on a number of different datasets, collected by various research groups using their own app and fingerprint readers. We also demonstrate that our matcher generalizes well to datasets which were not included during training. This cross-database evaluation solves a shortcoming of many existing studies which train and evaluate algorithms on different training and test splits of the same contact-contactless dataset. Furthermore, despite multiple evaluation datasets, we train only a single model for our evaluations, rather than fine-tuning individual models to fit a specific dataset.

Concretely, the contributions of our work are stated as:
\begin{enumerate}
    \item An end-to-end system, called C2CL, for contact-contactless fingerprint matching. C2CL is comprised of preprocessing (segmentation, enhancement, scaling, and deformation correction), feature extraction (minutiae and texture representations), and matching modules. Our preprocessing also benefits the Verifinger 12.0 commercial fingerprint SDK.
    
    \item A fully automated, preprocessing pipeline to map contactless fingerprints into the domain of contact-based fingerprints and a contact-contactless adaptation of DeepPrint~\cite{engelsma2019learning} for representation extraction. Our preprocessing and representation extraction is generalizable across multiple datasets and contactless capture devices.
    
    \item SOTA cross-matching verification and large-scale identification accuracy using C2CL on both publicly available contact-contactless matching datasets as well as on a completely sequestered dataset collected at Zhejiang University, China. Our evaluation includes the most diverse set of contactless fingerprint acquisition devices, yet we employ just a single trained model for evaluation.
    
    \item A smartphone contactless fingerprint capture app that was developed in-house for improved throughput and user-convenience. This app will be made available to the public to promote further research in this area\footnote{The project repository for the smartphone contactless fingerprint capture app is available at \url{https://github.com/ronny3050/FingerPhotos}.}.
    
    \item A new dataset of $9,888$ 2D contactless and corresponding contact-based fingerprint images from $206$ subjects ($2$ thumbs and $2$ index fingers per subject), which will be made available in the public domain to advance much needed research in this area\footnote{This dataset will be available to interested readers after this paper has been accepted.}.
\end{enumerate}

\begin{table*}
\caption{Summary of contact to contactless fingerprint recognition datasets used in this study.}
\label{tab:datasets}
\begin{tabularx}{\linewidth}{X | c | c | c | X | X}
\noalign{\hrule height 1.5pt}
          \multirow{1}{*}{\textbf{Dataset}} & \multirow{1}{*}{\textbf{\# Subjects}} & \multirow{1}{*}{\textbf{\# Unique}} & \multirow{1}{*}{\textbf{\# Images}} & \multirow{1}{*}{\textbf{Contactless Capture Device}} & \multirow{1}{*}{\textbf{Contact Capture Device}}\\[-0.3em]
          & & \textbf{Fingers} & \textbf{(Contactless / Contact)} & &\\
\noalign{\hrule height 1.0pt}
UWA Benchmark 3D Fingerprint Database, 2014~\cite{Zhou2014Benchmark}     & $150$      & $1,500$             & $3,000$ / $6,000$             & 3D Scanner (TBS S120E)  & CROSSMATCH Verifier 300 LC2.0\\
\hline
\multirow{5}{*}{ManTech Phase2, 2015~\cite{ericson2015evaluation}}   & \multirow{5}{*}{$496$}       & \multirow{5}{*}{$4,960$}              & \multirow{5}{*}{N/A / N/A$^*$} & \multirow{5}{\hsize}{AOS ANDI On-The-Go (OTG), MorphoTrak Finger-On-The-Fly (FOTF), IDair innerID on iPhone 4.}  & Cross Match Guardian R2, Cross Match SEEK Avenger, MorphoTrak MorphoIDent, MorphoTrust TouchPrint 5300, Northrop Grumman BioSled\\
\hline
PolyU Contactless 2D to Contact-based 2D Images Database, 2018~\cite{lin2018matching}     & N/A      & $336$               & $2,976$ / $2,976$             & Low-cost camera and lens (specific device not given) & URU 4000\\      
\hline
MSU Finger Photo and Slap Fingerprint Database, 2018~\cite{deb2018matching} & $309$ & $1,236$ & $2,472$ / $2,472$ & Xiaomi Redmi Note 4 smartphone & CrossMatch Guardian 200, SilkID (SLK20R)\\
\hline
IIT Bombay Touchless and Touch-Based Fingerprint Database, 2019~\cite{birajadar2019towards}  & N/A & $200$ & $800$ / $800$ & Lenovo Vibe k5 smartphone & eNBioScan-C1 (HFDU08)\\
\hline
\multirow{2}{*}{ISPFDv2, 2020~\cite{malhotra2020matching}}   & \multirow{2}{*}{$76$}       & \multirow{2}{*}{$304$}               & \multirow{2}{*}{$17,024$ / $2,432$}            & OnePlus One (OPO) and Micromax Canvas Knight smartphones  & \multirow{2}{*}{Secugen Hamster IV}\\
\hline
ZJU Finger Photo and Touch-based Fingerprint Database & $206$ & $824$ & $9,888$ / $9,888$ & HuaWei P20, Samsung s9+, and OnePlus 8 smartphones & URU 4500\\
\noalign{\hrule height 1.5pt}
\multicolumn{6}{l}{$^*$ The number of contact and contactless images acquired per finger varies for each device and the exact number is not provided.}\\
\end{tabularx}
\vspace{-1.5em}
\end{table*}

\section{Prior Work}

Prior studies on contact-contactless fingerprint matching primarily focus on only one of the sub-modules needed to obtain matching accuracy close to contact-contact based fingerprint matching systems (\textit{e.g.}, segmentation, distortion correction, or feature extraction only). These studies are categorized and discussed below.

\subsection{Segmentation}
The first challenge in contact-contactless matching is segmenting the relevant fingerprint region from the captured contactless fingerprint images. Malhotra \textit{et al}.~\cite{malhotra2020matching} proposed a combination of a saliency map and a skin-color map to segment the distal phalange (\textit{i.e.}, fingertip) of contactless fingerprint images in presence of varying background, illumination and resolution. Despite impressive results, the algorithm requires extensive hyperparameter tuning and still fails to accurately segment fingerprints in severe illumination conditions or noisy backgrounds. To alleviate these issues, we incorporate segmentation via an autoencoder trained to robustly segment the distal phalange of input contactless images.

\subsection{Enhancement}
One of the main challenges with contactless fingerprint images is the low ridge-valley contrast (Figure~\ref{fig:dataset_imgs}). The literature has addressed this in a number of different ways, including adaptive histogram equalization, Gabor filtering, median filtering, and sharpening by subtraction of the Gaussian blurred image from the captured image (\cite{dabouei2019deep, lin2019cnn-based,malhotra2020matching}). We also incorporate adaptive contrast enhancement in our work; however, one consideration that is lacking in existing approaches is the ridge inversion that occurs with Frustrated Total Internal Reflection (FTIR) optical imaging. In particular, the ridges and valleys of an FTIR fingerprint image will appear dark and light, respectively, while the opposite is true in contactless fingerprint images. Therefore, a binary inversion of the contactless fingerprint images is expected to improve the correspondence with their contact-based counterparts.

\subsection{Scaling}
After segmenting and enhancing a contactless fingerprint, the varying distances between fingers captured and the camera must be accounted for. In particular, since contact-based fingerprints are almost always captured at 500 pixels per inch (ppi), the contactless fingerprints need to be scaled to be as close to 500 ppi as possible. Previous studies have applied a fixed manual scaling, set for a specific dataset, or have employed contact-based fingerprint ridge frequency normalization algorithms that rely on accurate ridge extraction - which is often unreliable for contactless fingerprints. In contrast, we incorporate a spatial transformer network~\cite{jaderberg2015spatial} which has been trained to automatically normalize the resolution of the contactless fingerprints to match that of the 500 ppi contact images. This scaling is performed dynamically, \textit{i.e.} every input contactless fingerprint image is independently scaled.

\begin{figure*}[t]
    \centering
    \includegraphics[width=1.0\linewidth]{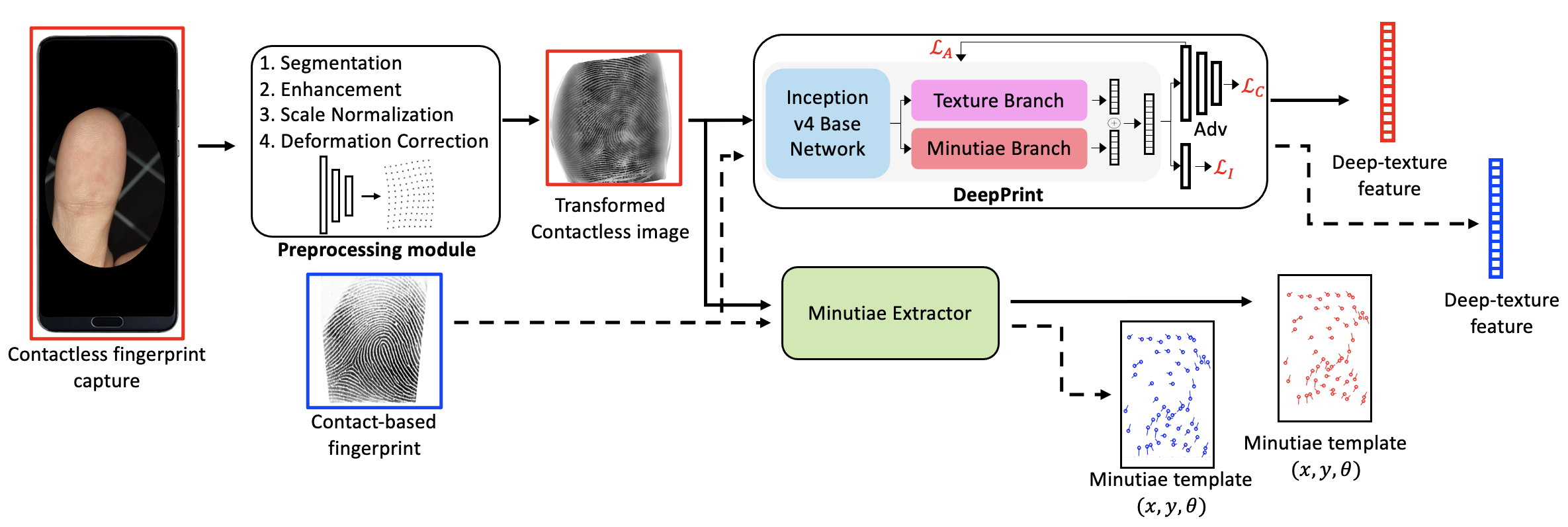}
    \caption{System architecture of C2CL. (a) A contactless fingerprint is captured and used as input to the preprocessing module, consisting of segmentation, enhancement, 500 ppi ridge frequency scaling, and deformation correction; (b) the transformed image output by the preprocessing module is fed to DeepPrint~\cite{engelsma2019learning}, which extracts a texture representation (shown in red). Without performing any additional preprocessing, the corresponding contact-based fingerprint is again fed to DeepPrint to extract a texture representation (shown in blue). Simultaneously, a minutiae representation is extracted using the Verifinger 12.0 SDK from both the contactless and contact-based fingerprint images.}
    \label{fig:3}
    \vspace{-1.5em}
\end{figure*}

\subsection{Distortion Correction}
A final preprocessing step for contact-contactless matching is non-linear distortion correction. To address this problem, \cite{lin2018matching} used thin-plate-spline (TPS) deformation correction models (previously applied for contact-contact matching~\cite{bazen2003fingerprint, dabouei2018fingerprint, ross2005fingerprint, ross2005deformable, senior2001improved, si2015detection}) using the alignment between minutiae annotations of corresponding contactless and contact fingerprints. A limitation is that the transformation is limited to one of six possible parameterizations. In a different study, Dabouei \textit{et al.}~\cite{dabouei2018fingerprint} train a spatial transformer to learn the distortion correction that is dynamically computed for each input image. In \cite{dabouei2018fingerprint}, a contact-based image is used as the reference for learning the distortion correction for a contactless image. However, we argue that this is not a reliable ground truth since the deformation varies among different contact-based fingerprint impressions. In our attempt to re-implement their algorithm, we found that this lack of a reliable and consistent ground truth makes training unstable, making it difficult to learn sound distortion parameters. In our work, rather than using the contact-based image as a reference, we use the match scores of our texture matcher as supervision for generating robust distortion correction. In other words, the distortion correction is optimized to maximize match scores between genuine contact-contactless fingerprint pairs. 

\subsection{Representation Extraction and Matching}
After preprocessing a contactless fingerprint image to lie within the same domain as a contact-based fingerprint, a discriminative representation must be extracted for matching. In the prior literature there are two main approaches to feature representation: (i) minutiae representation (\cite{dabouei2018fingerprint, lin2018matching}) and (ii) deep learning representation (\cite{lin2019cnn-based, malhotra2020matching}). Minutiae-based approaches rely on clever preprocessing and other techniques to improve the compatability of contactless fingerprint images for traditional contact-based minutiae extraction algorithms. On the other hand, deep learning approaches place less emphasis on preprocessing to manipulate the contactless fingerprints to improve correspondence with contact-based fingerprints, rather the responsibility is placed on the representation network to learn the correspondence despite the differences. For example, Lin and Kumar~\cite{lin2018contactless} and Dabouei \textit{et al.}~\cite{dabouei2018fingerprint} both apply a deformation correction to the contactless image to improve the minutiae correspondence. In contrast, the deep learning approach taken in \cite{lin2019cnn-based} applies very little preprocessing to the contactless image (just contrast enhancement and Gabor filtering) and leverages a Siamese CNN to extract features for matching. Similarly, Malhotra et al.~\cite{malhotra2020matching} utilize a deep scattering network to extract multi-scale and multi-directional feature representations.

In contrast to prior studies, our approach utilizes both a texture representation and a minutiae representation. Given the lower contrast and quality of contactless fingerprints (causing missing or spurious minutiae) and the non-linear distortion and scaling discrepancies between contact and contactless fingerprints (negatively impacting minutiae graph matching algorithms) a global texture representation is useful to improve the contact-contactless matching accuracy. We demonstrate this hypothesis empirically in the experimental results. 

\section{Methods}
Our matcher, C2CL, aims to improve contact to contactless fingerprint recognition through a multi-stage preprocessing algorithm and matching algorithm comprised of both a minutiae representation and a texture feature representation. The preprocessing is employed to minimize the domain-gap between the contactless fingerprints residing in a domain $\mathcal{D}_{cl}$ and contact-based fingerprints residing in another domain $\mathcal{D}_c$ and consists of segmentation, enhancement, ridge frequency scaling to $500$ ppi, and deformation correction through a learned spatial transformation network. After preprocessing, we extract deep-textural and minutiae representations (unordered, variable length sets $T = \{(x_1, y_1, \theta_1), ..., (x_n, y_n, \theta_n)\}$) for matching. The final match scores are obtained via a score-level fusion between the texture and minutiae matching scores.

\begin{figure}
    \centering
    \subfloat[]{\includegraphics[width=0.5\linewidth]{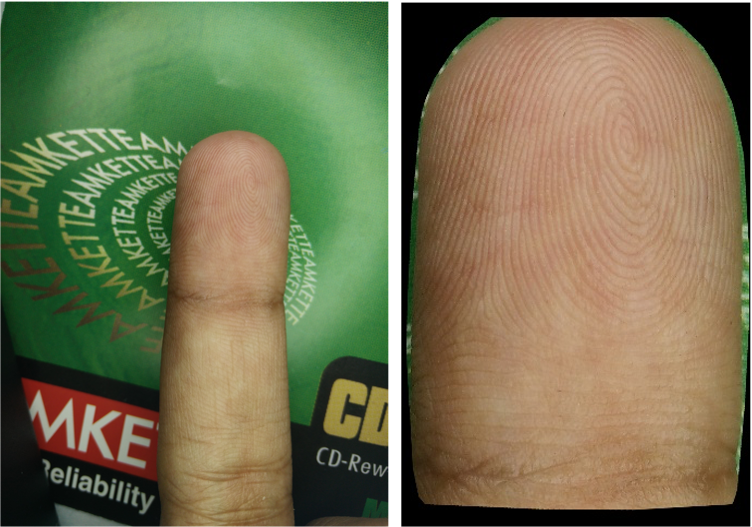}%
    \label{gen_seg_failues}}
    \subfloat[]{\includegraphics[width=0.5\linewidth]{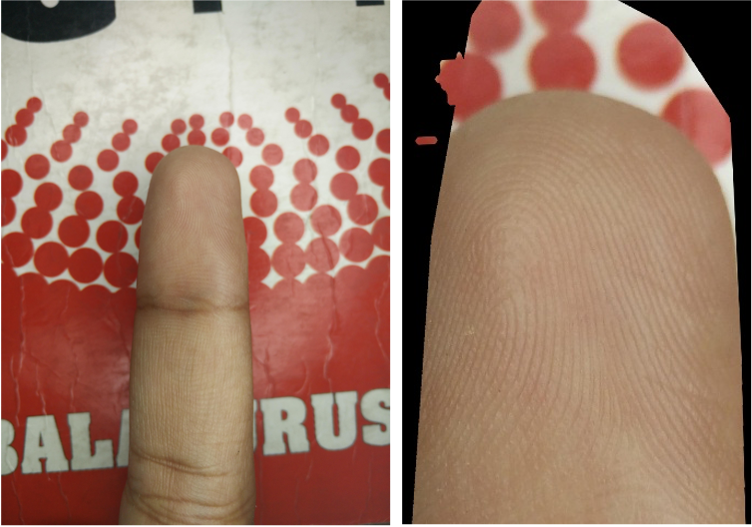}%
    \label{imp_seg_failures}}
    \caption{Example segmentation success (a) and failure (b) cases from images in the ISPFDv2 dataset using our segmentation algorithm. Sources of failure are presence of skin-like color tones in the background and varying skin complexion due to varying illumination.}
    \label{fig:seg_failures}
    \vspace{-1.5em}
\end{figure}

\begin{figure*}
    \centering
    \subfloat[]{\includegraphics[height=0.21\linewidth]{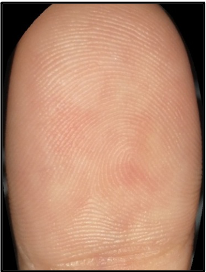}%
    \label{pre_B}}\hfill
    \subfloat[]{\includegraphics[height=0.21\linewidth]{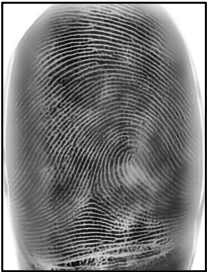}%
    \label{pre_C}}\hfill
    \subfloat[]{\includegraphics[height=0.21\linewidth]{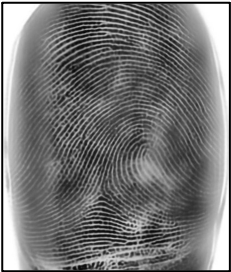}%
    \label{pre_D}}\hfill
    \subfloat[]{\includegraphics[height=0.21\linewidth]{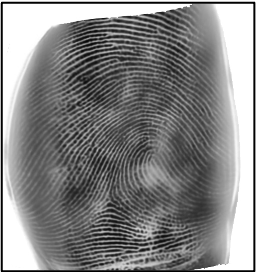}%
    \label{pre_E}}\hfill
    \subfloat[]{\includegraphics[height=0.21\linewidth]{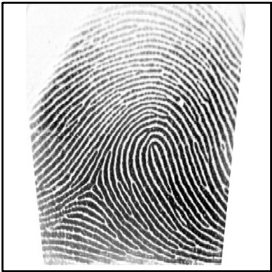}%
    \label{pre_F}}
    \caption{Illustration of our preprocessing pipeline including (a) segmentation, (b) enhancement, (c) scaling, and (d) warping. For reference, a corresponding contact-based fingerprint is shown in (e).}
    \label{fig:preprocessing}
    \vspace{-1.5em}
\end{figure*}

\subsection{Preprocessing}
Here we discuss the details of each stage of our preprocessing algorithm as illustrated in Figure~\ref{fig:preprocessing}.

\subsubsection{Segmentation}
Many contactless fingerprint datasets are unsegmented; for example, the ISPFDv2 dataset~\cite{malhotra2020matching} contains unsegmented, ($4,208\times3,120$) images with varying illumination, resolution, and background conditions. Thus, the first step in our preprocessing pipeline is to segment the distal phalange of the fingerprint using a U-net segmentation network~\cite{ronneberger2015u}. Our segmentation algorithm is a network $S(\cdot)$ which takes as input the unsegmented contactless fingerprint $I_{cl}$ of dimension ($m\times n$) and outputs a segmentation mask $\hat{M}$ $\in\{0, 1\}$ of dimension ($m\times n$). The obtained segmentation mask, $\hat{M}$, is element-wise multiplied with $I_{cl}$ to (i) crop out only the distal phalange of the contactless fingerprints and (ii) eliminate the remaining background to avoid detection of spurious minutiae in the later representation extraction stage. The segmented image $I'_{cl}$ is then resized to $480\times480$ by maintaining the aspect ratio with appropriate padding for further processing. 

For training $S(\cdot)$, we manually marked segmentation masks $M$ of the distal phalange of $496$ contactless fingerprints from the ISPFDv2 dataset\footnote{We used the open source Labelme segmentation tool found on GitHub~\cite{labelme2016}.}. Initially, $200$ images were randomly selected to have varying resolutions of either $5$MP, $8$MP, or $16$MP and another $200$ were selected with varying backgrounds and illumination. An additional $96$ images were specifically selected for their greater perceived difficulty, particularly images with skin tone backgrounds. The optimization function for training $S(\cdot$) is a pixel-wise binary cross-entropy loss between $\hat{M}$ and $M$ (Eq. \ref{eq:L_seg}).

\begin{multline}
    \mathcal{L}_{seg}(I_{cl},I_{c},M) = -\sum\limits_{i,j}[M_{i,j}\log(\hat{M_{i,j}}|I_{cl})\\+ (1-M_{i,j})\log(1-\hat{M_{i,j}}|I_{cl})]
    \label{eq:L_seg}
\end{multline}

\subsubsection{Enhancement}

Following segmentation, we apply a series of image enhancements $E(\cdot)$ to increase the contrast of the ridge-valley structure of the contactless images, including: (i) an adaptive histogram equalization to improve the ridge-valley contrast and (ii) pixel gray-level inversion to correct for the inversion of ridges between contact-based and contactless fingerprints. We also experimented with SOTA super resolution and de-blurring techniques, such as RDN~\cite{zhang2018residual}, to further improve the contactless image quality, but found only minimal matching accuracy improvements at the expense of significant additional computational cost.

\subsubsection{Distortion Correction and Scaling}
After segmenting and enhancing the contactless fingerprints, the non-linear distortions that separate the domains of contactless and contact-based fingerprints must be removed. In particular, this includes both a perspective distortion (caused by the varying distance of a finger from the camera) and a non-linear distortion (caused when the elastic human skin flattens against a platen).

To correct for these discrepancies, we train a spatial transformer network (STN)~\cite{jaderberg2015spatial} $T(\cdot)$ that takes as input a segmented, enhanced contactless image $I_{cl}^{e} = E(I'_{cl})$ and aligns the ridge structure to better match the corresponding contact-based image domain $\mathcal{D}_c$. The goal of the STN is two-fold: (i) an affine transformation $T_s(\cdot)$ to normalize the ridge frequency of the contactless images to match the $500$ ppi ridge spacing of the contact-based impressions and (ii) a TPS deformation warping $T_d(\cdot)$ of the contactless images to match the deformation present in contact-based images due to the elasticity of the human skin. 

Both $T_s(\cdot)$ and $T_d(\cdot)$ are comprised of a shared localization network $l(\cdot,w)$ and individual differentiable grid-samplers. Given an enhanced contactless fingerprint $I_{cl}^{e}$, $l(I_{cl}^{e},w)$ outputs the scale ($s$), rotation ($\theta$), and translation ($t_x, t_y$) of an affine transformation matrix $A_s$ (Eq. \ref{eq:A_s}) and a distortion field $\Theta$ which is characterized by a grid of $n\times n$ pixel displacements \{$(x_1,y_1)...(x_n,y_n)$\}. Subsequently, a scaled, warped image $I_{cl}^{w}$ is obtained via Equation~\ref{eq:warp}.

To learn the weights $w$ of the localization network such that $T_s(\cdot)$ and $T_d(\cdot)$ correctly scale the contactless fingerprints to 500 ppi, and unroll them into a contact-based fingerprint, we minimize the distance between DeepPrint representations extracted from genuine pairs of scaled, warped contactless fingerprints ($I_{cl}^{w}$) and contact-based fingerprints ($I_c$). In particular, let $f(\cdot)$ be a frozen DeepPrint network pretrained on contact-based fingerprints. Then, we can obtain a pair of 192D DeepPrint identity representations $R_{cl}$ and $R_{c}$ via $R_{cl} = f(I_{cl}^{w})$ and $R_c = f(I_{c})$. Our loss can then be computed from Equation~\ref{eq:loss_stn1}. By using the DeepPrint identity features extracted from contact-based fingerprint images to compute the loss, we are able to utilize the contact-based impressions as a ground truth of sorts. In particular, we are training our localization network to output better scalings and warpings such that the distortion and scale corrected contactless images have DeepPrint representations closer to their corresponding ``ground truth" contact-based image. 

We note that this approach has key differences to that which was proposed in~\cite{dabouei2019deep} where the distortion corrected contactless image (scale was not learned in~\cite{dabouei2019deep}) would be more directly compared to the ground-truth contact-based fingerprint via a cross-entropy loss between ``binarized" versions of $I_{cl}^{w}$ and $I_c$. We found that directly comparing the contactless and contact images via a cross entropy loss was quite difficult in practice since the ground truth contact image and the corresponding contactless image will have different rotations and translations separating them (even after scaling and distortion correction - resulting in a high loss value even if the scaling and distortion are correct). Furthermore, the contact-based image itself varies based upon the pressure applied during the acquisition, environmental conditions, sensor model, \textit{etc.}, meaning that directly using the contact-based image as ground truth is unreliable. In contrast, since DeepPrint has been trained to be invariant to pressure, environmental conditions, and sensor model, our ground truth (DeepPrint representations from contact-based images) will remain stable across different contact-based impressions. In short, unlike~\cite{dabouei2019deep}, we learn both distortion correction \textbf{and} scaling correction simultaneously, and we use the DeepPrint identity loss to stabilize training of $T(\cdot)$ and to enable predictions of warpings and scalings which better improve matching accuracy.

\begin{equation}
    A_s = \begin{bmatrix} s\cos(\theta)&-s\sin(\theta)&t_x\\s\sin(\theta)&s\cos(\theta)&t_y \end{bmatrix}
    \label{eq:A_s}
\end{equation}

\begin{equation}
    I_{cl}^{w} = T(I_{cl}^e;A_s,\Theta) = T_d(T_s(I_{cl}^e, A_s), \Theta)
    \label{eq:warp}
\end{equation}

\begin{equation}
    \mathcal{L}_{STN} = \lVert R_{cl} - R_{c}\rVert^{2}_{2}
    \label{eq:loss_stn1}
\end{equation}

\begin{table*}
\caption{Number of contactless and contact fingerprint images used in training each component of C2CL (\# contactless / \# contact).}
\label{tab:datasets_used}
\begin{tabularx}{\linewidth}{@{\extracolsep{\fill}}l M M M}
\noalign{\hrule height 1.5pt}
 \multirow{2}{*}{\textbf{Dataset}} & \textbf{Segmentation} & \textbf{Deformation Correction \& Scaling} & \textbf{DeepPrint}\\
 & \mathbf{S(\cdot)} & \mathbf{T(\cdot)} & \mathbf{f(\cdot)} \\
\noalign{\hrule height 1.0pt}
UWA Benchmark 3D Fingerprint Database~\cite{Zhou2014Benchmark}  & 0 / 0 & 0 / 0 & 1,000 / 2,000\\
\hline
ManTech Phase2, 2015~\cite{ericson2015evaluation} & 0 / 0 & 0 / 0 & 21,352 / 28,574\\
\hline
PolyU Contactless 2D to Contact-based 2D Images Database~\cite{lin2018matching} & 0 / 0 & 1,920 / 1,920 & 1,920 / 1,920\\
\hline
MSU Finger Photo and Slap Fingerprint Database~\cite{deb2018matching}  & 0 / 0 & 2,472 / 2,472 & 2,472 / 2,472\\
\hline
IIT Bombay Touchless and Touch-based Fingerprint Database~\cite{birajadar2019towards} & 0 / 0 & 800 / 800 & 800 / 800\\
\hline
ISPFDv2~\cite{malhotra2020matching}  & 496 / 0 & 8,400 / 1,200 & 8,400 / 1,200\\
\hline
ZJU Finger Photo and Touch-based Fingerprint Database & 0 / 0 & 0 / 0 & 0 / 0\\
\noalign{\hrule height 1.0pt}
\textbf{Total} & \mathbf{496 / 0} & \mathbf{13,592 / 6,392} & \mathbf{35,944 / 36,966}\\
\noalign{\hrule height 1.5pt}
\end{tabularx}
\vspace{-1.5em}
\end{table*}

\subsection{Representation Extraction}
After performing all of the aforementioned preprocessing steps, we enter the second major stage of our contact-contactless matcher, namely the representation extraction stage. Our representation extraction algorithm extracts both a textural representation (using a CNN) and a minutiae-set. Scores are computed using both of these representations and then fused together using a sum score fusion.

\subsubsection{Texture Representation}
To extract our textural representation, we fine-tune the DeepPrint network proposed by Engelsma \textit{et al.} in~\cite{engelsma2019learning} on a training partition of the publicly available datasets which we aggregated (Table~\ref{tab:datasets_used}). Unlike the deep networks used in~\cite{lin2019cnn-based} and \cite{malhotra2020matching} for extraction of textural representations, DeepPrint is a deep-network which has been specifically designed for fingerprint representation extraction via a built-in alignment module and minutiae domain knowledge. Therefore, in this work, we seek to adopt DeepPrint for contact-contactless fingerprint matching. As is common practice in the machine learning and computer vision communities, we are utilizing a pretrained DeepPrint network to warm start our model, which has been shown to improve over random initialization for many applications; for example, in fingerprint spoof detection~\cite{nogueira2016fingerprint}.

Formally, DeepPrint is a network $f(\cdot)$ with parameters $w$ that takes as input a fingerprint image $I$ and outputs a fixed-length fingerprint representation $R$ (which encodes the textural related features). During training, DeepPrint is guided to encode features related to fingerprint minutiae via a multi-task learning objective including: (i) a cross-entropy loss on both the minutiae branch identity classification probability $\hat{y}_1$ and texture branch identity classification probability $\hat{y}_2$ (Eq. \ref{eq:1}), (ii) minimize the intra-class variance of class $y$ via a center loss between the predicted minutiae feature vector $R_1$ and its mean feature vector $R^y_1$ and the predicted texture feature vector $R_2$ and its mean feature vector $R^y_2$ (where $R_1$ concatenated with $R_2$ form the full representation $R$), and (iii) a mean squared error loss on the predicted minutiae maps $\hat{H}$ output by DeepPrint's minutiae branch and ground truth minutiae maps H (Eq. \ref{eq:3}). These losses are combined to form the DeepPrint identity loss, $\mathcal{L}_{ID}$ (Eq. \ref{eq:4}), where $\lambda_1 = 1$, $\lambda_2 = 0.00125$, $\lambda_3 = 0.095$ are set empirically.

\begin{equation}
    \mathcal{L}_1(I,y) = -\log(\hat{y}^{j=y}_1|I,w) - \log(\hat{y}^{j=y}_2|I,w)]
    \label{eq:1}
\end{equation}

\begin{equation}
    \mathcal{L}_2(I,y) = \lVert R_1 - \Bar{R}^y_{1} \rVert^{2}_{2} + \lVert R_2 - \Bar{R}^y_{2}\rVert^{2}_{2}
    \label{eq:2}
\end{equation}

\begin{equation}
    \mathcal{L}_3(I,H) = \sum\limits_{j,k,l}(\hat{H}_{j,k,l} - H_{j,k,l})^2
    \label{eq:3}
\end{equation}

\newcommand{\argmin}{\mathop{\mathrm{argmin}}\limits}
\begin{multline}
    \mathcal{L}_{ID}(I,y,H) = \argmin_w \sum\limits_{i=1}^{N}[\lambda_1\mathcal{L}_1(I^i,y^i) + \lambda_2\mathcal{L}_2(I^i,y) \\ + \lambda_3\mathcal{L}_3(I^i,H^i)]
    \label{eq:4}
\end{multline}

Due to the differences in resolution, illumination, and backgrounds observed between different datasets of contactless fingerprint images, generalization to images captured on unseen cameras becomes critical. The problem of cross-sensor generalization in fingerprint biometrics (\textit{e.g.}, optical reader to capacitive reader), of which contact to contactless matching is an extreme example, has been noted in the literature~\cite{lugini2013interoperability, ross2004biometric, alonso2006sensor, alshehri2018large}, with many previous works aimed at improving the interoperability~\cite{marasco2013minimizing, jang2007improving, ross2006calibration}. Motivated by the recent work employing adversarial learning to cross-sensor generalization of fingerprint spoof detection~\cite{grosz2020fingerprint}, we incorporate an adversarial loss to encourage robustness of DeepPrint to differences between acquisition devices. The adversarial loss $\mathcal{L}_A$ is defined as the cross-entropy on the output of an adversary network $q(\cdot, \theta_A)$ across $C$ classes of sensors, where the adversarial ground truth $y'$ is assigned equal probabilities across these $C$ classes (Eq. \ref{eq:5}). The adversarial loss $\mathcal{L}_A$ and identity loss $\mathcal{L}_{ID}$ form the overall loss function $\mathcal{L}_D$ used to train DeepPrint (Eq. \ref{eq:6}), where $\lambda_4 = 0.1$ is empirically selected. The adversary network, $q(\cdot, \theta_A)$, is a two layer fully connected network, with weights $\theta_A$, that predicts the probability of the class of input device used to capture each image, \textit{i.e.}, minimizes the cross-entropy of the predicted device and the ground truth device label $y$ (Eq. \ref{eq:7}). Intuitively, if DeepPrint learns to fool the adversary, it has learned to encode identifying features which are independent of the acquisition device or camera.

\begin{equation}
    \mathcal{L}_A(I,y') = -\sum\limits_{c=1}^{C}y_c'\log q_{A}(y_c|f(I;w);\theta_{A})
    \label{eq:5}
\end{equation}

\begin{multline}
    \mathcal{L}_D(I,y,H,y') = \argmin_w \sum\limits_{i=1}^{N}[\mathcal{L}_{ID}(I,y,H) \\+ \lambda_4\mathcal{L}_A(I^i,y'^i)]
    \label{eq:6}
\end{multline}

\begin{equation}
    \mathcal{L}_C(I,y_c) = -y_c\log q_{A}(y_c|f(I;w);\theta_{A})]
    \label{eq:7}
\end{equation}

In addition to the adversarial loss, we also increased the DeepPrint representation dimensionality from the original 192D to 512D and added perspective distortion and scaling augmentations during training. In an ablation study (Table~\ref{tab:deepprint_ablation}), we show how each of our DeepPrint modifications (fine-tuning, adversarial loss, perspective and scaling augmentations, and dimensionality change) improves the contact-contactless fingerprint matching performance. 

\subsubsection{Minutiae Representation}
Finally, after extracting a textural representation with our modified DeepPrint network, we extract a minutiae-based representation from our preprocessed contactless fingerprints with the Verifinger 12.0 SDK.

\subsection{Matching}
Following feature extraction, from which we obtain texture representations ($R_t^c$, $R_t^{cl}$) and Verifinger minutiae representations ($R_m^c$, $R_m^{cl}$) for a given pair of contact and contactless fingerprint images ($I_c$, $I_{cl}$), we compute a final match score as a weighted fusion of the individual scores computed between ($R_t^c$, $R_t^{cl}$) and ($R_m^c$, $R_m^{cl}$). Concretely, let $s_t$ denote the similarity score between ($R_t^c$, $R_t^{cl}$) and $s_m$ denote the similarity score between ($R_m^c$, $R_m^{cl}$), then the final similarity score is computed from a sum score fusion shown in Equation~\ref{eq:score_fusion}. For our implementation, $w_t$ = $w_m$ = $0.5$ was selected empirically.

\begin{equation}
    s = w_ts_t + w_ms_m
    \label{eq:score_fusion}
\end{equation}

\section{Experiments}
In this section, we give details on various experimental evaluations to determine the effectiveness of C2CL for contact to contactless fingerprint matching. We employed various publicly available datasets for the evaluation of our algorithms, as well as a new database of contactless and corresponding contact-based fingerprints which was collected using our mobile-app in coordination with Zhejiang University (ZJU).

\subsection{Datasets}
Table~\ref{tab:datasets} gives a detailed description of the publicly available datasets for contact to contactless matching used in this study and Figure~\ref{fig:dataset_imgs} shows some example images from these datasets. For comparison with previous studies, we use the same train/test split of the PolyU dataset that was used in \cite{lin2019cnn-based}, which consists of $160$ fingers for training with $12$ impressions each and the remaining $160$ fingers for testing with $6$ impressions each. Similarily, we split the UWA Benchmark 3D dataset into $500$ training fingers and $1,000$ unique test fingers. Furthermore, following the protocol of Malhotra \textit{et al.}~\cite{malhotra2020matching}, we split the ISPFDv2 dataset evenly into $50\%$ train and $50\%$ test subjects. Finally, we captured and sequestered a new dataset of contactless fingerprints and contact-based fingerprint images in coordination with ZJU for a cross-database evaluation (\textit{e.g.}, not seen during training) to demonstrate generalizability of our algorithm. The cross-database evaluation is much more stringent than existing approaches which only train/test on different partitions of the same dataset. Indeed, the cross-database evaluation is a much better measure of how C2CL would perform in the real world.

The ZJU Finger Photo and Touch-based Fingerprint Database contains a total of $206$ subjects, with $12$ contactless images and $12$ contact-based impressions per finger. The thumb and index fingers of both hands were collected for each subject, giving a total of $9,888$ contactless and contact-based images each. The contactless images were captured using three commodity smartphones: HuaWei P20, Samsung s9+, and OnePlus 8, whereas the contact-based fingerprint impressions were captured on a URU 4500 optical-based scanner at 512 ppi. An Android fingerphoto capture app was developed to improve the ease and efficiency of the data collection. To initiate the capturing process, a user or operator enters the transaction ID for the user and uses an on screen viewing window to help guide and capture the fingerprint image. Furthermore, a counter displayed on the screen keeps track of subsequent captures to streamline the data collection process.

\subsection{Implementation Details}
All the deep learning components (segmentation network, deformation correction and scaling network, and DeepPrint) are implemented using the Tensorflow deep learning framework. Each network is trained independently and information regarding how many of the contactless and contact fingerprint images from each of the datasets used in training each component of our algorithm is given in Table~\ref{tab:datasets_used}.

\subsubsection{Segmentation Network}
A total of $496$ contactless fingerprint images from the ISPFDv2 were manually labeled with segmentation masks outlining the distal phalange were used for training. Input images were down sampled to $256\times256$ during training to reduce the time to convergence, which occurred around $100,000$ iterations using stochastic gradient descent (SGD) with a learning rate of $1e^{-3}$ and a batch size of $8$ on a single NVIDIA GeForce RTX 2080 Ti GPU. During inference, the contactless fingerprint images are resized to $256\times256$ and resulting segmentation masks are upsampled back to the original resolution. Due to limited number of manually marked images, we employed random rotations, translations, and brightness augmentations to avoid over-fitting. Additionally, we incorporated random resizing of input training images within the range [$128\times128$, $384\times384$] to encourage robustness to varying resolution between capture devices.

\subsubsection{Deformation Correction and Scaling Network}
The pretrained DeepPrint model in~\cite{engelsma2019learning} was used to provide supervision of our spatial transformation network in line with Eq~\ref{eq:loss_stn1}. The motivation for using a network pretrained on contact-based fingerprints, rather than our new finetuned model on contactless fingerprints, is that the goal of our transformation network is to transform the contactless fingerprint images to better resemble their contact-based counterparts. Thus, a supervisory network trained on solely contact-based fingerprint images is more suitable for this purpose. The architectural details of our STN localization network are given in Table~\ref{tab:STN}. For our implementation, we set the number of sampling points for the distortion grid to n = $4\times4$. Data augmentations of random rotations, translations, brightness adjustments, and perspective distortions were employed to avoid over-fitting. This network was trained for $25,000$ iterations using an Adam optimizer with a learning rate of $1e^{-6}$ and a batch size of $16$ on a single NVIDIA GeForce RTX 2080 Ti GPU.

\begin{table}
\caption{Deformation Correction and Scaling Spatial Transformation Network Architecture, $T(\cdot)$.}
\label{tab:STN}
\begin{tabularx}{\linewidth}{X || M || M}
\noalign{\hrule height 1.5pt}
\textbf{Layer}    & \textbf{\#Filters, Filter Size, Stride} & \textbf{Output Dim.} \\
\noalign{\hrule height 1.0pt}
\hline
0. Input            &          0,0,0       & 480\times480\times1\\
\hline
1. Convolution      &  32, 3\times3, 2 & 240\times240\times32\\
\hline
2. Convolution      &  64, 3\times3, 2 & 120\times120\times64\\
\hline
3. Convolution      &  128, 3\times3, 2 & 60\times60\times128\\
\hline
4. Convolution      &  256, 3\times3, 2 & 30\times30\times256\\
\hline
5. Max Pool   &  256,6\times4, 2 & 13\times14\times256\\
\hline
6. Dense            &  - & 1024\\
\hline
7. Dense            & -  & 2\times n_0 + 4\\
\noalign{\hrule height 1.5pt}
\multicolumn{3}{p{0.95\linewidth}}{The final dense layer contains output neurons for a $2\times n_0$ grid of $n_0 = n\times n$ pixel displacements and 4 neurons for the affine transformation matrix ($s$, $\theta$, $t_x$ and $t_y$.). In our implementation, $n=4$.}\\
\end{tabularx}
\vspace{-1.5em}
\end{table}

\subsubsection{DeepPrint}
The DeepPrint network was trained on two NVIDIA GeForce RTX 2080 Ti GPUs with an RMSProp optimizer, learning rate of $0.01$, and a batch size of $16$. The added adversary network, which was trained in step with DeepPrint, also utilized an RMSProp optimizer with a learning rate of $0.01$. A small validation set was partitioned from the DeepPrint fine-tuning data outlined in Table~\ref{tab:datasets_used} to stop the training (which occurred at 73,000 steps). Lastly, random rotation, translation, brightness, cropping, and perspective distortion augmentations were utilized during training.

\subsection{Evaluation Protocol}
To evaluate the cross-matching performance of our algorithms, we conduct both verification (1:1) and identification (1:N) experiments. For verification, we report the Receiver Operating Characteristic (ROC) curves at specific operating points and equal error rates (EER). Note that we report the True Acceptance Rate (TAR) at a False Acceptance Rate (FAR) of $0.01\%$, which is a stricter threshold than is currently reported in the literature, and which is also a threshold expected for field deployment. For the search experiments, the rank-one search accuracy is given against an augmented large scale gallery of $1.1$ contact million fingerprints taken from an operational forensics database~\cite{yoon2015longitudinal}. This is a much larger gallery than has previously been evaluated against in the literature and is again more indicative of what C2CL would face in the real world. Finally, we present ablation results on each significant component of our proposed system.

\begin{table*}[t]
\caption{Verification performance of C2CL.}
\label{tab:verification_results}
\begin{tabularx}{\linewidth}{@{\extracolsep{\fill}}l || M | M || M | M || M | M || M}
\noalign{\hrule height 1.5pt}
\multirow{2}{*}{\textbf{Dataset}} & \multicolumn{2}{c||}{\textbf{Verifinger 12.0}} & \multicolumn{2}{c||}{\textbf{DeepPrint}} & \multicolumn{2}{c||}{\textbf{DeepPrint + Verifinger 12.0}} & \textbf{Previous SOTA}\\\cline{2-8}
 & \textbf{EER (\%)} & \textbf{TAR @ FAR=0.01\%} & \textbf{EER (\%)} & \textbf{TAR @ FAR=0.01\%} & \textbf{EER (\%)} & \textbf{TAR @ FAR=0.01\%} & \textbf{EER (\%)}\\
\noalign{\hrule height 1.0pt}
\text{PolyU} & 0.46 & 97.20 & 2.37 & 72.07 & 0.30 & 97.74 & 7.93~\cite{lin2019cnn-based}\\
\hline
\text{UWA} & 6.81 & 92.56 & 5.29 & 83.40 & 0.72 & 98.30 & 7.11~\cite{lin2019cnn-based}\\
\hline
\text{ISPFDv2} & 1.46 & 96.02 & 2.33 & 84.33 & 1.20 & 96.67 & 3.40^\ddagger~\cite{malhotra2020matching}\\
\hline
\text{ZJU}$^\dagger$ & 0.79 & 96.86 & 2.08 & 86.42 & 0.62 & 97.56 & \text{N/A}\\
\noalign{\hrule height 1.5pt}
\multicolumn{8}{l}{$^\ddagger$ \cite{malhotra2020matching} reports results on the ISPFDv2 dataset per individual capture condition; 3.40 is the average EER across these data splits.}\\
\multicolumn{8}{l}{$^\dagger$ Cross-database evaluation, \textit{i.e.}, not seen during training.}\\
\end{tabularx}
\vspace{-1.5em}
\end{table*}

\subsection{Verification Experiments}
The verification experiments are conducted in a manner consistent with previous approaches to facilitate a fair comparison. In particular, (i) the PolyU testing dataset yields $5,760$ ($160\times6\times6$) genuine scores and $915,840$ ($160\times159\times6\times6$) imposter scores, (ii) the UWA Benchmark 3D dataset yields $8,000$ ($1,000\times4\times2$) genuine and $7,992,000$ ($1,000\times999\times4\times2$) imposter scores, (iii) the ISPFDv2 dataset (which is split into $7$ different capture variations)\footnote{The $7$ scenarios consist of different background, illumination, and resolution variations (\textit{e.g.}, white background \& indoor lighting, white background \& outdoor lighting, natural background \& indoor lighting, natural background \& outdoor lighting, $5$MP resolution, $8$MP resolution, and $16$MP resolution. For our evaluation, we combine each of these into a single dataset.} yields $68,096$ ($(152\times8\times8)\times7)$ genuine and $10,282,496$ ($(152\times151\times8\times8)\times7$) imposter scores, and (iv) the ZJU dataset yields $118,656$ ($824\times12\times12$) genuine and $97,653,888$ ($824\times823\times12\times12$) imposter scores. Due to the very high number of possible imposter scores for ZJU, we limit the number of imposter scores computed to only include the first impression of each imposter fingerprint. This process results in $678,152$ imposter scores out of the possible $97,653,888$ scores. It is assumed for all experiments that the contactless fingerprints and contact-based impressions are the probe and enrollment images, respectively.

Table~\ref{tab:verification_results} provides the Equal Error Rate (EER) and TAR @ FAR=$0.01\%$ of C2CL on the different datasets. For comparison with previous methods, rather than implement the relevant SOTA approaches that have been proposed and risk under representing those methods, we directly compare our approach to the results reported in each of the respective papers. In terms of EER, our method outperforms all the previous approaches in the verification setting. Not only does our individual performance of the minutiae and textural representations alone exceed that of the previous SOTA methods (in particular, even if we remove Verifinger, we still beat SOTA in all cases), the fusion performance attains matching accuracy (EER = $0.30\%-1.20\%$), which is much closer to contact-contact fingerprint matching~\cite{dorizzi2009fingerprint}. Even in the most challenging cross-database evaluation (ZJU), C2CL attains competitive performance with contact-contact matching - demonstrating the generalizability of C2CL to unseen datasets. Note that we report the TAR @ FAR=$0.01\%$ only for C2CL since most of the prior approaches only report EER and none report TAR @ FAR =$0.01\%$.

Different from previous approaches, which train individual models on a train/test split for each evaluation dataset, we have trained just a single model for our evaluation across four different datasets. This protocol is actually more challenging than finetuning for each individual evaluation dataset. This is because despite having a smaller number of training samples, higher verification performance can more easily be achieved by individually trained models. To support this claim, we have finetuned an additional model on just the PolyU dataset using the same train/test split specified in \cite{lin2019cnn-based} and recorded the verification performance in Table~\ref{tab:polyu_only}. We observe that our accuracy improves on PolyU from 2.37\% EER for the model trained on our full combination of training datasets to 1.90\% EER for the model trained on just PolyU; however, because of the lower performance on the other three datasets, we can see that this model is indeed over-fit to PolyU.

\begin{table}
    \caption{DeepPrint verification performance when finetuned on only PolyU.}
    \label{tab:polyu_only}
    \begin{tabularx}{1.0\linewidth}{ l | M | M }
    \noalign{\hrule height 1.5pt}
    \textbf{Dataset} & \textbf{EER (\%)} & \textbf{TAR (\%) @ FAR=0.01\%} \\
    \hline
    PolyU & $1.90$ & $74.62$ \\
    \hline
    UWA & $8.35$ & $35.42$ \\
    \hline
    ISPFDv2 & $3.87$ & $57.10$ \\
    \hline
    ZJU & $2.99$ & $68.99$ \\
    \noalign{\hrule height 1.5pt}
    \end{tabularx}
\end{table}

\subsubsection{Ablation study}
We present an ablation study (Table~\ref{tab:ablation}) to fully understand the contribution of the main components of our algorithm; namely, segmentation, enhancement, 500 ppi frequency scaling, and TPS deformation correction. From the ablation, we notice there is a substantial improvement in both EER and TAR @ FAR=$0.01\%$ just from incorporating proper enhancement of the contactless images. In most cases, there is almost a $50\%$ reduction in EER from including both contrast enhancement and binary pixel inversion. For brevity, not shown in the table is the individual contribution of inverting the ridges of the contactless images aside from contrast enhancement. For reference, the EER of DeepPrint on ZJU warped images with only contrast enhancement is $2.49\%$. This is in comparison to the EER of $2.08\%$ on the warped images with both contrast enhancement and pixel inversion.

Furthermore, we observe that for the smartphone captured contactless fingerprints in the ISPFDv2 and ZJU datasets, there is a dramatic performance jump when incorporating our $500$ ppi scaling network. Additionally, there is another noticeable improvement when incorporating the deformation correction branch of our STN, most notably for the ISPFDv2 dataset. Since the ZJU dataset contains equal numbers of thumb and index fingers, where the majority of our training datasets contain mostly non-thumb fingers, we observed that the deformation correction is less beneficial on average for the ZJU dataset compared to ISPFDv2. In fact, from Table~\ref{tab:multifinger}, we see that the EER of just index fingers of ZJU is noticeably lower than the EER on thumbs.

To investigate whether the lower performance on thumbs is a limitation of the available training data or whether thumbs require a different distortion correction from non-thumbs, we retrained separate warping models on thumb data only and non-thumb data only. The test results for the ZJU dataset are in Table~\ref{tab:diff_warping}. A couple of observations: i.) The performance (TAR @ FAR=0.01\%) is highest for the model trained on both thumbs and non-thumbs, ii.) the model trained on non-thumbs performs slightly worse when applied to the test set of thumbs in the ZJU dataset, which indicates that the warping required for thumbs may be slightly different, and iii.) the performance of the thumb only model decreases on both thumbs and non-thumbs due to the limited number of thumb training examples.

\begin{table}[t]
\caption{Ablation study of C2CL using only Verifinger 12.0 for matching$^*$. $S$ = segmentation, $E$ = enhancement $T_s$ = scaling, $T_d$ = deformation correction.}
\label{tab:ablation}
\begin{tabularx}{\linewidth}{l || l | l | l | l || M | M}
\noalign{\hrule height 1.5pt}
\multirow{2}{*}{\textbf{Dataset}} & \multicolumn{4}{c||}{\textbf{Modules}} & \multicolumn{2}{c}{\textbf{Overall} ($\%$)}\\\cline{2-7}
& $\mathbf{S}$ & $\mathbf{E}$ & $\mathbf{T_s}$ & $\mathbf{T_d}$ & \textbf{EER} & \textbf{TAR @ FAR = 0.01\%} \\
\noalign{\hrule height 1.0pt}
\multirow{4}{*}{PolyU} & \checkmark & & & & 0.86 & 93.19 \\
& \checkmark & \checkmark & & & 0.45 & 96.96\\
& \checkmark & \checkmark & \checkmark & & 0.48 & 96.44\\
& \checkmark & \checkmark & \checkmark & \checkmark & 0.46 & 97.20\\
\hline
\multirow{2}{*}{UWA$^\ddagger$} & \checkmark & & & & 6.62 & 91.05\\
& \checkmark & \checkmark & & & 6.81 & 92.56 \\
\hline
\multirow{4}{*}{ISPFDv2} & \checkmark & & & & 13.76 & 23.93\\
& \checkmark & \checkmark & & & 7.83 & 38.53\\
& \checkmark & \checkmark & \checkmark & & 2.02 & 93.3\\
& \checkmark & \checkmark & \checkmark & \checkmark & 1.46 & 96.02\\
\hline
\multirow{4}{*}{ZJU$^\dagger$} & \checkmark & & & & 3.35 & 82.8\\
& \checkmark & \checkmark & & & 1.88 & 89.9\\
& \checkmark & \checkmark & \checkmark & & 0.9 & 96.97\\
& \checkmark & \checkmark & \checkmark & \checkmark & 0.79 & 96.86\\
\noalign{\hrule height 1.5pt}
\multicolumn{7}{l}{$^*$ Ablation results for DeepPrint are not shown since only a single}\\
\multicolumn{7}{l}{model was trained on the final $E$+$S$+$T_s$+$T_d$ images.}\\
\multicolumn{7}{l}{$^\ddagger$ We do not apply our STN here since these images are captured with}\\
\multicolumn{7}{l}{a 3D scanner and are already unrolled and at a resolution of $500$ ppi.}\\
\multicolumn{7}{l}{$^\dagger$ Cross-database evaluation, \textit{i.e.}, not seen during training.}\\
\end{tabularx}
\end{table}

\begin{table}[t]
\small
\caption{DeepPrint Ablation Study}
\label{tab:deepprint_ablation}
\begin{tabularx}{\linewidth}{L{0.4\linewidth} | M }
\noalign{\hrule height 1.5pt}
\textbf{Method} & \textbf{ZJU EER (\%)} \\
\noalign{\hrule height 1.0pt}
DeepPrint~\cite{engelsma2019learning} & 4.07\\
\hline
+ finetune & 2.68\\
\hline
+ 512D & 2.64\\
\hline
+ Augmentations & 2.35\\
\hline
+ Adversarial Loss & 2.08\\
\noalign{\hrule height 1.5pt}
\multicolumn{2}{l}{$^*$ Each row adds on to the previous row.}\\
\end{tabularx}
\end{table}

\subsubsection{Multi-finger fusion verification}
The final set of verification experiments is to investigate the effects of finger position and multiple finger fusion in the verification accuracy for the ZJU dataset. Table \ref{tab:multifinger}, shows the individual performance per finger position and the fusion of multiple fingers; namely, thumb only, index only, fusion of right thumb and right index, fusion of left thumb and left index, and four finger fusion. The motivation for considering fusion of the thumb and index on each hand is that from a usability standpoint, a user may be able to use their dominant hand when capturing their own fingerprints. Notably, when fusing multiple fingers (\textit{e.g.}, right index and left index), we obtain nearly perfect accuracy.

\begin{table}[t]
\small
\centering
\caption{Multi-finger fusion verification results of the proposed matcher on the ZJU dataset.}
\label{tab:multifinger}
\begin{tabularx}{\linewidth}{X | M | M}
\noalign{\hrule height 1.5pt}
\textbf{Finger Type} & \textbf{EER (\%)} & \textbf{TAR (\%) @ FAR = 0.01\%}\\
\noalign{\hrule height 1.0pt}
Thumb & 0.95 & 95.89\\
\hline
Index & 0.48 & 98.31\\
\hline
LT + LI & 0.00 & 99.77\\
\hline
RT + RI & 0.00 & 99.74\\
\hline
RT + LT & 0.00 & 99.80\\
\hline
RI + LI & 0.00 & 99.89\\
\noalign{\hrule height 1.5pt}
\end{tabularx}
\end{table}

\begin{table}
\caption{Applying separate warping modules for thumbs vs. non-thumbs (TAR (\%) @ FAR=0.01\%).}
\label{tab:diff_warping}
\begin{tabularx}{1.0\linewidth}{ X | X | X | X }
\noalign{\hrule height 1.5pt}
\textbf{Method} & \textbf{Trained on \newline thumbs and \newline non-thumbs} & \textbf{Trained on \newline non-thumbs} & \textbf{Trained on \newline thumbs} \\
\hline
ZJU \newline non-thumbs & $92.22$ & $92.24$ & $91.66$ \\
\hline
ZJU thumbs & $80.66$ & $77.50$ & $76.80$ \\
\hline
ZJU all & $86.42$ & $84.80$ & $84.46$ \\
\noalign{\hrule height 1.5pt}
\end{tabularx}
\end{table}

\begin{table}
\caption{Improvement in minutiae correspondence without and with warping correction on ZJU dataset.}
\label{tab:minu_corr}
\begin{tabularx}{1.0\linewidth}{ X | X | X | X | X }
\noalign{\hrule height 1.5pt}
 & \textbf{Avg. Number of \newline Paired Minutiae} & \textbf{Avg. Number of \newline Missing Minutiae} & \textbf{Avg. Number of \newline Spurious Minutiae} & \textbf{Goodness Index}~\cite{ratha1995adaptive} \\
\hline
Without warping & $28.06$ & $67.95$ & $71.13$ & -0.0167  \\
\hline
With \newline warping & $\mathbf{30.11}$ & $\mathbf{65.00}$ & $\mathbf{69.20}$ & $\mathbf{-0.0157}$ \\
\noalign{\hrule height 1.5pt}
\end{tabularx}
\end{table}

\subsection{Search Experiments}
For the identification (or search) experiments, we utilize the first impressions of both the contactless and contact-based fingerprints of the ZJU dataset. The contact-based fingerprints are placed in the gallery which is augmented with 1.1 million fingerprint images from an operational forensic database~\cite{yoon2015longitudinal}. The contactless fingerprint images serve as the probes. We note that our 1.1 million augmented gallery is significantly larger than any of the existing galleries used to evaluate contact-contactless fingerprint search, and is more indicative of the real world use-case of cross fingerprint matching (\textit{e.g.}, in a National ID system like Aadhaar where a large gallery of contact-based fingerprints is already enrolled and used for de-duplication). 

We evaluate 3 different search algorithms on the ZJU augmented gallery: (i) Verifinger 1:N search, (ii) search via our DeepPrint texture matcher (scores $s_t$ from Eq.~\ref{eq:score_fusion} are computed between a given preprocessed, contactless probe, and all 1.1 million contact-based fingerprints in the gallery), and (iii) a two-stage search algorithm~\cite{engelsma2019learning} where the DeepPrint texture scores are first used to retrieve the top-500 candidates, followed by a reordering using the 1:1 minutiae matching scores ($s_m$ from Eq.~\ref{eq:score_fusion}) from Verifinger. The advantage of the two-stage search scheme is that it balances both speed and accuracy by utilizing the matching speed of DeepPrint to locate the first list of 500 candidates and the accuracy of Verifinger to further refine this list.

From Table~\ref{tab:identification_results}, we observe that Verifinger outperforms DeepPrint stand-alone but at a search time against 1.1 million that is quite slow in comparison to DeepPrint. This motivates combining both approaches into the aforementioned two-stage search algorithm which outperforms Verifinger at Rank-1 and reduces the search time by 50 seconds. In short, our two stage search algorithm obtains high levels of search accuracy on a large-scale gallery at a significant search time savings.

\begin{table}[h]
\small
\caption{Search performance of the proposed matcher on the ZJU dataset with a gallery of $1.1$ million.}
\label{tab:identification_results}
\begin{tabularx}{\linewidth}{X | M | M | M | M | M}
\noalign{\hrule height 1.5pt}
\multirow{2}{*}{\textbf{Method}} & \multicolumn{4}{c|}{\textbf{Rank (\%)}} & \textbf{Search}\\\cline{2-5}
 & \textbf{1} & \textbf{10} & \textbf{100} & \textbf{500} & \textbf{Time (s)}\\
\noalign{\hrule height 1.0pt}
DeepPrint & 83.56 & 93.06 & 95.86 & 97.08 & 0.4 \\
\hline
Verifinger 12.0 & 95.25 & 96.47 & 96.95 & 97.20 & 60.1 \\
\hline
DeepPrint + Verifinger 12.0 & 95.49 & 96.10 & 96.95 & 97.08 & 10.5 \\
\noalign{\hrule height 1.5pt}
\multicolumn{6}{p{0.95\linewidth}}{DeepPrint + Verifinger 12.0 refers to indexing the top-500 candidates with DeepPrint and then re-sorting those 500 candidates using a fusion of the Verifinger and DeepPrint score.}\\
\end{tabularx}
\end{table}

\subsection{Segmentation Evaluation}
A successful segmentation algorithm for contactless fingerprint images must not only reliably detect the distal phalange of the contactless fingerprint, but also be robust to varying illumination, background, and resolution that is expected to occur in highly unconstrained capture environments. The method by Malhotra \textit{et al.}~\cite{malhotra2020matching} performed well on the ISDFPDv2 dataset using certain hyperparameters that were fit to this particular dataset; however, the authors did not evaluate it on unseen datasets. In contrast, our algorithm requires no hyperparameter tuning and still performs well across a variety of different evaluation datasets, both seen and unseen. Table~\ref{tab:IOU} gives a comparison on the unseen ZJU dataset between our method and our implementation of the baseline approach of Malhotra \textit{et al.}, which was trained on the ISPFDv2 dataset. For this evaluation, we manually marked the first contactless fingerprint image of each unique finger in the ZJU dataset with ground truth segmentation masks of the distal phalange, and then computed the Intersection Over Union (IOU) metric between the predicted segmentation masks of our algorithm and our implementation of the benchmark algorithm in \cite{malhotra2020matching}. Our method does not require any hyper-parameter tuning and still achieves higher IOU compared to \cite{malhotra2020matching}.

A qualitative analysis of our segmentation network (see Figure~\ref{fig:seg_failures}) shows our algorithm is robust to varying illumination, background, and resolution and generalizes across multiple datasets of contactless fingerprints. However, as seen in Figure~\ref{fig:seg_failures} (b), the network may still fail in extremely challenging background and illumination settings. An additional consideration, which is of importance for real-time deployment, is the processing speed of the segmentation network. Our segmentation algorithm is extremely fast compared to existing methods - requiring just $12.6$ms to segment a ($900\times1200$) resolution image. In contrast, our parallel implemenation of the baseline approach of Malhotra \textit{et al.} requires $3$s per image.

\begin{table}
\small
\caption{Intersection Over Union (IOU) for segmentation $S(\cdot)$.}
\label{tab:IOU}
\begin{tabularx}{\linewidth}{L{0.4\linewidth} | M }
\noalign{\hrule height 1.5pt}
\textbf{Method} & \textbf{IOU} \\
\noalign{\hrule height 1.0pt}
Baseline~\cite{malhotra2020matching} & 0.747\\
\hline
\textbf{Proposed} & 0.899\\
\noalign{\hrule height 1.5pt}
\end{tabularx}
\end{table}

\section{Discussion}
Despite the low error rates achieved across each dataset, there are many factors that complicate the cross-matching performance and lead to both type I (false rejects) and type II (false accepts) errors. Many of the type I and type II errors are attributed to a failure to correctly segment and scale only the distal phalange of the input contactless fingerprint. Incorrect segmentation can lead the large amounts of the image containing background rather than the relevant fingerprint region. Other errors can be attributed to the inherent low-contrast of the contactless fingerprints, despite any effort of contrast or resolution enhancement. The only way to mitigate these types of failures is to include a quality assurance algorithm at the point of capture of the contactless fingerprint images. Lastly, minimal overlap in the fingerprint ridge structure between genuine probe and gallery fingerprint images is the cause of many false rejections, whereas very similar ridge structure between imposter fingerprint pairs leads to a number of false accepts. This challenge is present in contact-contact matching; however, is exaggerated in C2CL because of the unconstrained pose variance of the finger in 3D space.

The potential for greater variance in the capture conditions when capturing contactless fingerprint images necessitates more robust preprocessing to reliably match contactless fingerprints. Thus, performance will likely be markedly lower in unconstrained scenarios compared to highly controlled capture environments that employ dedicated hardware for the image acquisition, such as the PolyU and UWA datasets. However, C2CL has pushed the SOTA forward both in matching more unconstrained fingerphotos and the more constrained dedicated-device captured contactless fingerprints. Additionally, one might consider acquiring multiple image views of the same finger to build a complete 3D model of the finger to guide the preprocessing stage; however, this would add additional computational costs and latency to the acquisition process. Furthermore, in some capture scenarios, certainly the setup employed by our capture app, this process may be ergonomically challenging for the user.

\begin{figure}
    \centering
    \subfloat[]{\includegraphics[width=1.0\linewidth]{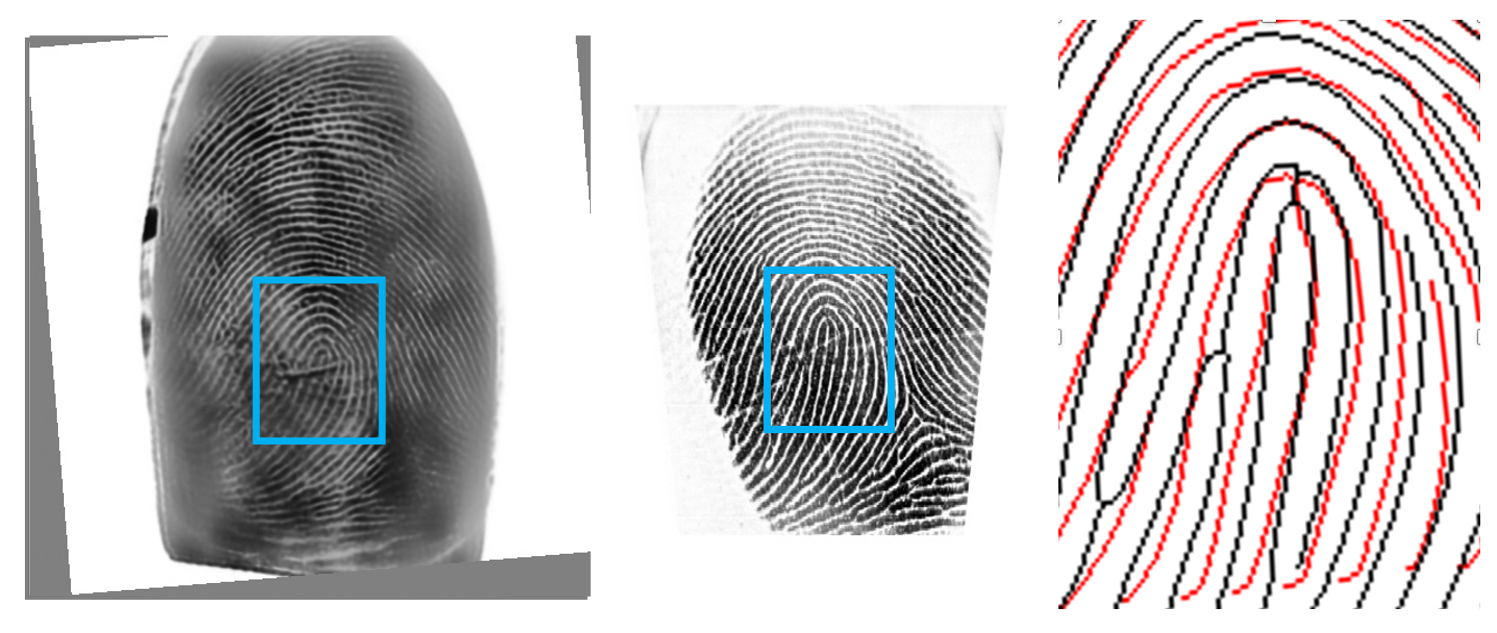}%
    \label{toprow_overlay}}\\
    \subfloat[]{\includegraphics[width=1.0\linewidth]{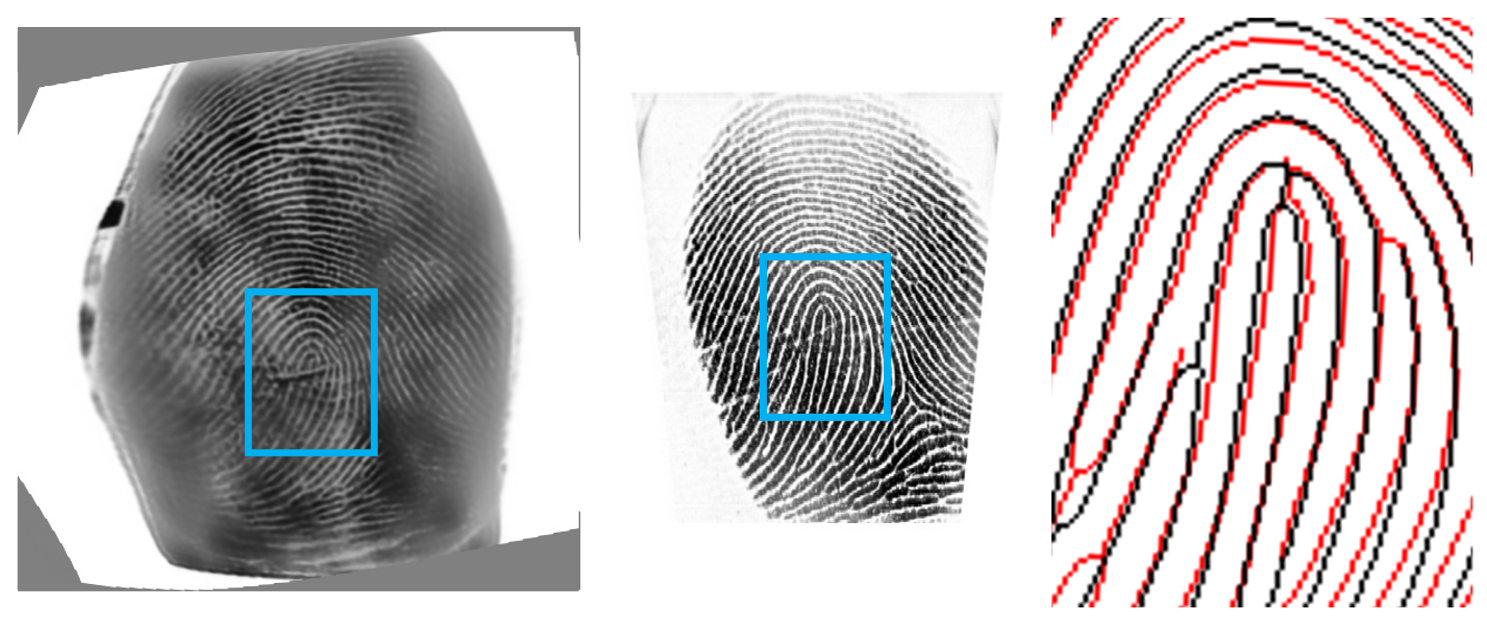}%
    \label{bottomrow_overlay}}
    \caption{Comparison of ridge overlap (a) with and (b) without the unwarping module. Use of the unwarping module results in better ridge alignment between contactless and contact-based images.}
    \label{fig:ridgeOverlay}
\end{figure}

As highlighted in the ablation study of Table~\ref{tab:verification_results}, most of the improvement in interoperability between contactless and contact-based fingerprints is due to appropriate 500 ppi scaling of the contactless prints; however, incorporating a deformation correction module is also shown, with statistical significance\footnote{The Mann-Whitney rank test~\cite{delong1988comparing} was used to compute the statistical significance between the ROC curves of $S$+$E$+$T_s$ and $S$+$E$+$T_s$+$T_d$. For all four datasets, the p value is smaller than $0.05$, indicating that the difference is statistically significant to reject the hypothesis that the two curves are similar with a confidence of $95\%$.}, to further improve the compatibility. Figure~\ref{fig:ridgeOverlay} aims to highlight this fact through an overlay of the fingerprint ridge structure of one pair of corresponding contact and contactless fingerprints before and after applying the deformation correction. Additionally,  Table~\ref{tab:minu_corr} shows the average number of paired minutiae, missing minutiae, spurious minutiae, and Goodness Index~\cite{ratha1995adaptive} without and with the warping correction on the ZJU dataset. The GI, ranging from -1 to 3, is a combined measure of paired, missing, and spurious minutiae. The warping module improved the GI by $5.99\%$. Thus, the improved alignment indubitably leads to better minutiae-based and texture-based matching, as verified by our experiments.

Lastly, in order to utilize a large CNN, such as DeepPrint, for the task of contact-contactless fingerprint matching, we leveraged a large dataset~\cite{yoon2015longitudinal} from a related domain of contact-contact fingerprint matching to pretrain our DeepPrint network. Since this dataset is not currently publicly available, we have repeated the verification experiments when pretraining DeepPrint on the publicly available NIST N2N dataset~\cite{fiumara2018nist} (see Table~\ref{tab:N2N_pretraining}). Due to the smaller dataset, we experience a slight degradation in the DeepPrint performance on some of the evaluation datasets; however, further data augmentation and incorporation of other publicly available datasets can be used to improve the performance.

\begin{table*}[t]
\caption{DeepPrint performance pretrained with NIST N2N~\cite{fiumara2018nist} dataset vs. Longitudinal dataset referenced in Yoon and Jain~\cite{yoon2015longitudinal}.}
\label{tab:N2N_pretraining}
\begin{tabularx}{\linewidth}{l || X | X || X | X}
\noalign{\hrule height 1.5pt}
\multirow{2}{*}{\textbf{Dataset}} & \multicolumn{2}{c||}{\textbf{Pretrained on NIST N2N Dataset (publicly available)}} & \multicolumn{2}{c}{\textbf{Pretrained on Longitudinal Dataset}}\\\cline{2-5}
& \textbf{EER (\%)} & \textbf{TAR (\%) @ FAR = 0.01\%} & \textbf{EER (\%)} & \textbf{TAR (\%) @ FAR = 0.01\%} \\
\noalign{\hrule height 1.0pt}
PolyU & 2.04 & 71.30 & 2.37 & 72.07 \\
\hline
UWA & 5.62 & 56.99 & 5.29 & 83.40 \\
\hline
ISPFDv2 & 2.60 & 81.83 & 2.33 & 84.33 \\
\hline
ZJU$^\dagger$ & 3.09 & 77.92 & 2.08 & 86.42 \\
\noalign{\hrule height 1.5pt}
\multicolumn{5}{l}{$^\dagger$ Cross-database evaluation, \textit{i.e.}, not seen during training.}\\
\end{tabularx}
\end{table*}

\section{Computational Efficiency}
Our system architecture consists of a variety of deep networks (segmentation network, deformation correction and scaling STN, DeepPrint CNN feature extractor) and a minutiae feature extractor. The inference speeds of the segmentation network, STN, and DeepPrint are approximately $12.6$ms, $6.2$ms, and $26.3$ms using a single NVIDIA GeForce RTX 2080 Ti GPU and $143.8$ms, $19.5$ms, and $120.2$ms on an Intel Core i7-8700X CPU @ 3.70GHz, respectively. The Verifinger 12.0 feature extractor requires $600$ms on an Intel Core i7-8700X. In total, the inference speed of the end-to-end network is $\approx643.6$ms with an NVIDIA GeForce RTX 2080 Ti GPU or $\approx883.5$ms on an Intel Core i7-8700X CPU.

The deep network components of our algorithm are capable of very fast inference per input image; however, the system as a whole consumes a large amount of memory ($400$ MB). To fit into a resource constrained environment, such as a mobile phone, further optimization to the system architecture can easily be implemented with very little, if any, performance drop. First, the intermediate step of generating a scaled image prior to the deformation correction is not required for deployment and was just included for the ablation study. Instead, we can remove the affine transformation layer of our STN and directly scale and warp the input images in one step. As it stands, the main components of the algorithm, DeepPrint and Verfinger, require $\approx1$s and $\approx1.2$s on a mobile phone (Google Pixel 2), respectively. Thus, the inference time is estimated to be $\approx2$ seconds. However, to further boost the speed, rather than rely on a COTS system for minutiae extraction and matching, we can directly use the minutiae sets output by DeepPrint and a computationally efficient minutiae matcher to obtain the minutiae match scores, such as MSU's Latent AFIS Matcher~\cite{cao2019end}. Porting the model with these optimizations to a mobile phone remains as a point of future work.

\section{Conclusion and Future Work}
We have presented an end-to-end system for matching contactless fingerprints (\textit{i.e.}, finger photos) to contact-based fingerprint impressions that significantly pushes the SOTA in contact-contactless fingerprint matching closer to contact-contact fingerprint matching. In particular, our contact to contactless matcher achieves less than $1\%$ EER across multiple datasets employing a variety of contactless and contact-based acquisition devices with varying background, illumination, and resolution settings. Critical to the success of our system is our extensive preprocessing pipeline consisting of segmentation, contrast enhancement, 500 ppi scale normalization, deformation correction, and our adaptation of DeepPrint for contact-contactless matching. Our cross-database evaluations and large-scale search experiments are more rigorous evaluations than what is reported in the open literature, and it enables us to confidently demonstrate a step toward a contact-contactless fingerprint matcher that is comparable to SOTA contact-contact fingerprint matching accuracy.

\section*{Acknowledgment}
This material is based upon work supported by the Center for Identification Technology Research and the National Science Foundation under Grant No. 1841517. The authors would like to thank Debayan Deb for his help in developing the mobile phone contactless fingerprint capture application, Dr. Eryun Liu's research group at Zhejiang University for overseeing the data collection effort for this project, and the various research groups who have shared their datasets that were used in this study.

\ifCLASSOPTIONcaptionsoff
  \newpage
\fi

\bibliographystyle{IEEEtran}
\bibliography{citationsTIFS.bib}
\begin{IEEEbiography}[{\includegraphics[width=1in,height=1.25in,clip,keepaspectratio]{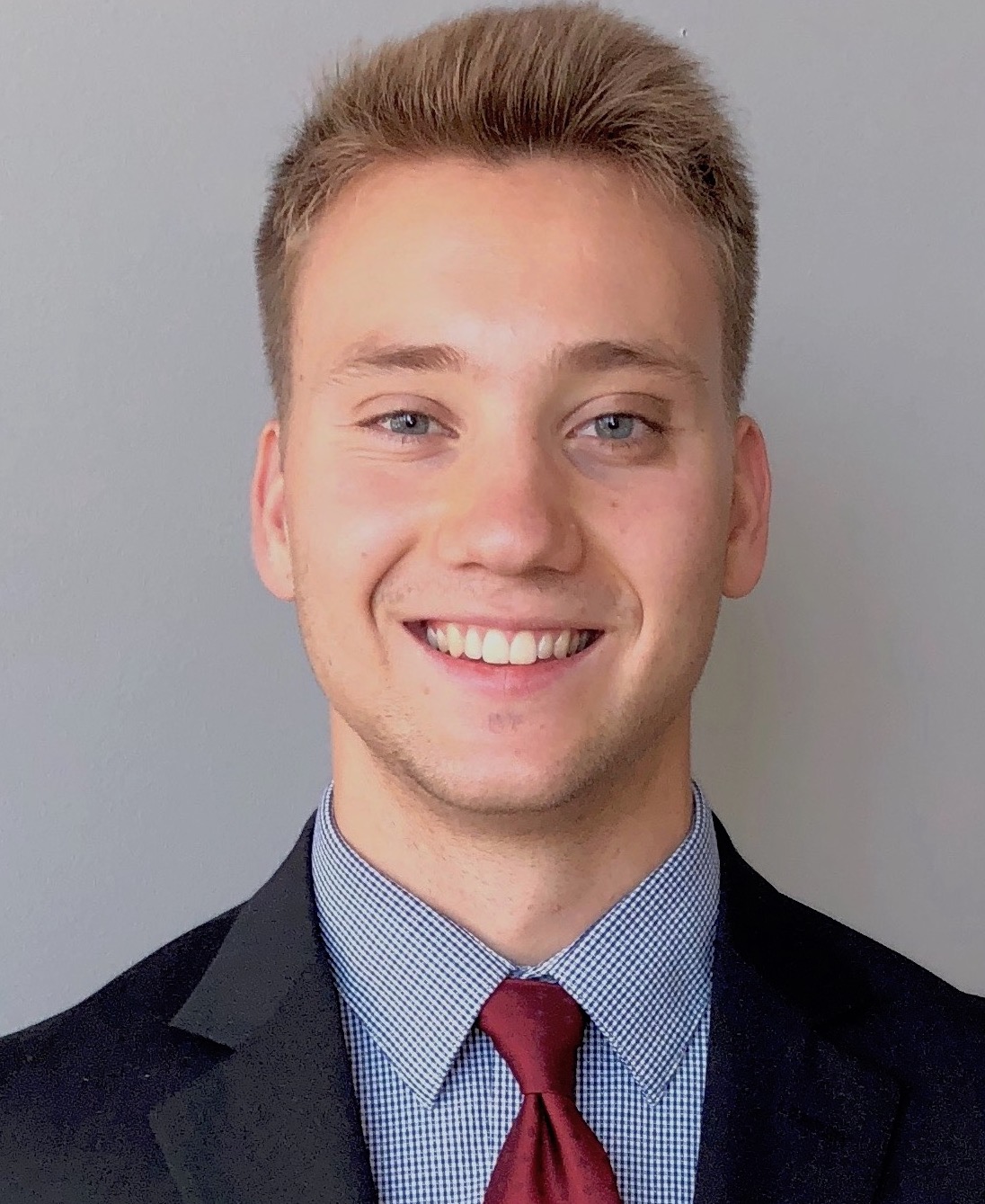}}]{Steven A. Grosz}
received his B.S. degree with highest honors in Electrical Engineering from Michigan State University, East Lansing, Michigan, in 2019. He is currently a doctoral student in the Department of Computer Science and Engineering at Michigan State University. His primary research interests are in the areas of machine learning and computer vision with applications in biometrics.
\end{IEEEbiography}
\begin{IEEEbiography}[{\includegraphics[width=1in,height=1.25in,clip,keepaspectratio]{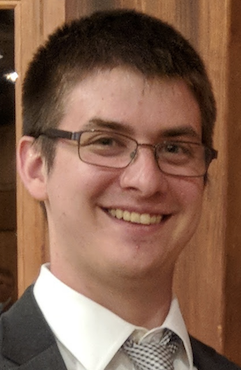}}]{Joshua J. Engelsma}
graduated magna cum
laude with a B.S. degree in computer science from Grand Valley State University, Allendale, Michigan, in 2016. He is currently working towards a PhD degree in the Department of Computer Science and Engineering at Michigan State University, East Lansing, Michigan. His research interests include pattern recognition, computer vision, and image processing with applications in biometrics. He won best paper award at the 2019 IEEE ICB, and the 2020 MSU College of Engineering Fitch Beach Award.
\end{IEEEbiography}
\begin{IEEEbiography}[{\includegraphics[width=1in,height=1.25in,clip,keepaspectratio]{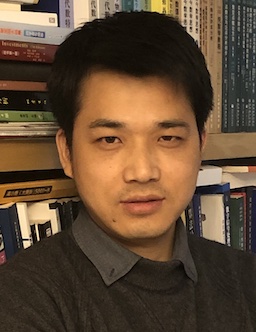}}]{Eryun Liu}
received his Bachelor degree in electronic information science and technology from Xidian University, Xi'an, Shaanxi China, in 2006 and Ph.D. degree in Pattern Recognition and Intelligence System from the same university in 2011. He was affiliated with Xidian University as an assistant professor from 2011 to 2013. He was a Post Doctoral Fellow in the Department of Computer Science \& Engineering, Michigan State University, East Lansing, USA. He is currently affiliated with the Department of Information Science \& Electronic Engineering (ISEE), Zhejiang University, Hangzhou, China, as an associate professor. His research interests include biometric recognition, point pattern matching and information retrieval, with a focus on fingerprint and palmprint recognition.
\end{IEEEbiography}
\begin{IEEEbiography}[{\includegraphics[width=1in,height=1.25in,clip,keepaspectratio]{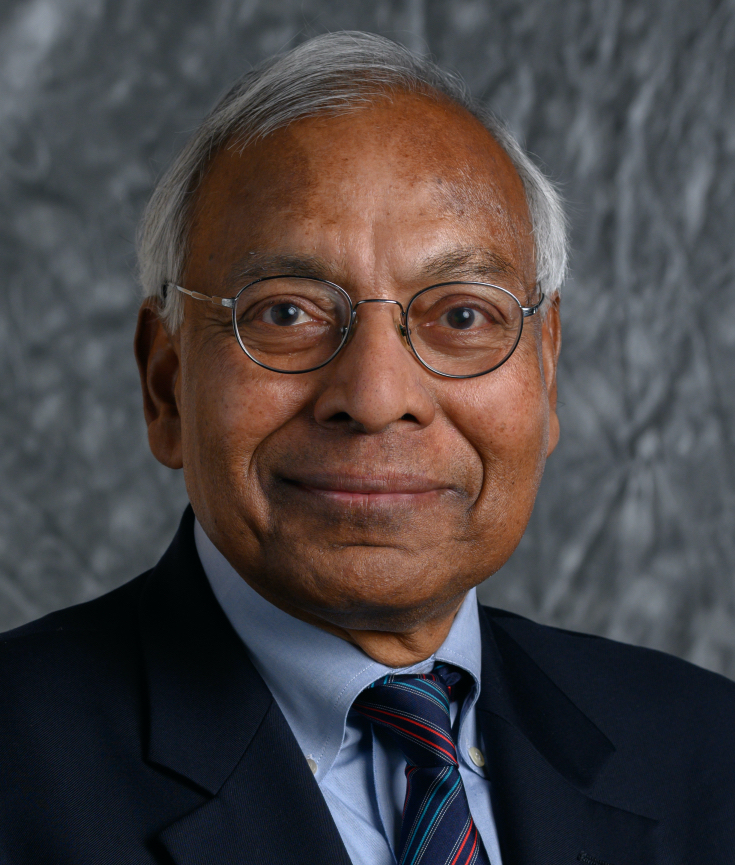}}]{Anil K. Jain}
is a University distinguished professor in the Department of Computer Science and Engineering at Michigan State University. He served as the editor-in-chief of the IEEE Transactions on Pattern Analysis and Machine Intelligence and was a member of the United States Defense Science Board. He has received Fulbright, Guggenheim, Alexander von Humboldt, and IAPR King Sun Fu awards. He is a member of the National Academy of Engineering and foreign fellow of the Indian National Academy of Engineering and Chinese Academy of Sciences.
\end{IEEEbiography}

\end{document}